\documentclass{vmax-preprint}

\usetikzlibrary{arrows.meta,calc,positioning}
\tcbuselibrary{listings}

\newtcblisting{promptbox}[1][]{%
  listing only,
  breakable,
  colback=gray!5,
  colframe=gray!50,
  arc=2pt,
  boxrule=0.5pt,
  left=4pt, right=4pt, top=4pt, bottom=4pt,
  before skip=6pt, after skip=6pt,
  fonttitle=\bfseries\small,
  listing options={
    basicstyle=\ttfamily\footnotesize,
    breaklines=true,
    breakatwhitespace=true,
    columns=fullflexible,
  },
  #1
}

\title{Breaking the Solver Bottleneck: Training Task Generators at the Learnable Frontier}

\author[1]{Lorenz Wolf}
\author[2]{Connor Watts}
\author[1]{Roger Creus Castanyer}
\author[1]{Geoffrey Bradway}
\author[1]{Maxwill Lin}
\author[1]{Augustine N. Mavor-Parker}
\author[1]{Matthew Daborn-Sargent}

\affiliation[1]{Vmax}
\affiliation[2]{Goodfire AI}

\abstract{%
The limiting resource for training agents via reinforcement learning (RL) is increasingly frontier task supply: valid, solvable tasks just difficult enough to train the current model. As reasoning and agentic models improve, fixed task distributions saturate, while naive synthetic generation yields tasks that are trivial, impossible, or ill-posed. Training a task generator with RL to optimize validity and learnability can address this bottleneck, but direct optimization requires repeated solver rollouts per candidate. For software-engineering (SWE) tasks, a single rollout can take tens of minutes; solver-in-the-loop generator training is intractable. We introduce PROPEL, a solver-amortized framework for training task generators at the targeted solve rate. PROPEL trains a lightweight activation probe on a one-time labeled corpus of generated tasks and solver outcomes. The probe predicts target-solver pass rate from a frozen generator reference model and serves as a proxy for solve rate during generator optimization, reducing generator evaluation to a single forward pass. Across math, code, and software-engineering at multiple model scales, PROPEL shifts generation toward the targeted solve rate: for coding, tasks generated at the learnable frontier increase from $10.1\% \rightarrow 20.0\%$ for a \texttt{Qwen2.5-3B-Instruct} solver and from $5.3\% \rightarrow 12.6\%$ for a \texttt{Qwen2.5-7B-Instruct} solver. For SWE, PROPEL increases the share of generations at the targeted solve rate from $9.8\%$ to $19.6\%$ for \texttt{Qwen3.5-27B} on repositories not seen during training of probe and generator.
}

\date{\today}
\correspondence{\email{\{lorenz, augustine, matthew\}@vmax.ai}}

\begin{document}
\maketitle



\section{Introduction}

%
\begin{figure}[t]
\centering
\resizebox{\linewidth}{!}{%
\begin{tikzpicture}[
  x=0.92cm,y=0.92cm,
  >=Latex,
  font=\sffamily\scriptsize,
  box/.style={
    draw=black!70,
    fill=white,
    rounded corners=1.8pt,
    line width=0.75pt,
    align=center,
    inner sep=4pt
  },
  probebox/.style={
    box,
    fill=green!10
  },
  flow/.style={
    ->,
    draw=black!80,
    line width=0.85pt
  },
  frozen/.style={
    ->,
    draw=black!65,
    dashed,
    line width=0.85pt
  },
  reward/.style={
    ->,
    draw=red!75!black,
    line width=1.05pt
  },
  stage/.style={
    font=\sffamily\bfseries\small,
    align=center
  },
  smalllabel/.style={
    font=\sffamily\tiny,
    align=center
  },
  redlabel/.style={
    font=\sffamily\bfseries\tiny,
    text=red!75!black,
    align=center
  },
  captionbox/.style={
    font=\sffamily\tiny,
    align=center,
    rounded corners=1.6pt,
    inner sep=3pt
  },
  pics/doc/.style={
    code={
      \draw[black!70,line width=0.5pt,fill=white]
        (0,0) -- (0,0.36) -- (0.20,0.36) -- (0.30,0.26) -- (0.30,0) -- cycle;
      \draw[black!70,line width=0.5pt]
        (0.20,0.36) -- (0.20,0.26) -- (0.30,0.26);
      \draw[black!55,line width=0.35pt] (0.06,0.27) -- (0.20,0.27);
      \draw[black!55,line width=0.35pt] (0.06,0.18) -- (0.20,0.18);
      \draw[black!55,line width=0.35pt] (0.06,0.09) -- (0.18,0.09);
    }
  }
]

\draw[draw=blue!35, fill=blue!5, rounded corners=2pt, line width=0.7pt]
  (0.00,0.00) rectangle (5.65,5.65);
\draw[draw=green!35, fill=green!5, rounded corners=2pt, line width=0.7pt]
  (5.85,0.00) rectangle (10.55,5.65);
\draw[draw=orange!40, fill=orange!6, rounded corners=2pt, line width=0.7pt]
  (10.75,0.00) rectangle (15.95,5.65);
\draw[draw=black!25, fill=black!3, rounded corners=2pt, line width=0.7pt]
  (16.15,0.00) rectangle (18.90,5.65);

\node[stage,text=blue!70!black]   at (2.825,5.30) {1\quad Data Collection};
\node[stage,text=green!55!black]  at (8.20,5.30)  {2\quad Probe Training};
\node[stage,text=orange!85!black] at (13.35,5.30) {3\quad RL Training};
\node[stage,text=black!65]        at (17.52,5.30) {4\quad Evaluation};


\node[box, minimum width=1.30cm, minimum height=0.85cm] (base1) at (0.95,4.40)
  {Base\\Generator\\[-0.4mm]{\tiny frozen}};

\node[smalllabel] at (3.20,4.85) {Generated tasks};
\pic at (2.55,4.22) {doc};
\pic at (2.95,4.22) {doc};
\pic at (3.35,4.22) {doc};
\pic at (3.75,4.22) {doc};
\coordinate (doc1west)  at (2.55,4.40);
\coordinate (docsouth)  at (3.30,4.22);
\coordinate (docseast1) at (4.05,4.40);

\node[smalllabel] at (4.95,4.85) {Activations};
\node[anchor=center, font=\sffamily] (acts1) at (4.95,4.40) {$h$};

\node[box, minimum width=1.80cm, minimum height=1.30cm] (solver1) at (1.20,2.55)
  {Solver Model\\[1.0mm]
   {\tiny $k{=}8$ trials per task}\\[-0.2mm]
   {\tiny slow \& expensive}};

\node[font=\sffamily\bfseries\tiny, anchor=west]
  at (2.82,3.40) {Difficulty labels};

\node[smalllabel,anchor=west] at (2.82,3.08)
  {$0/8$~{\color{red!75!black}$\boldsymbol{\times}$}~{\tiny too hard}};
\node[smalllabel,anchor=west] at (2.82,2.74)
  {$1/8$~{\color{green!55!black}$\checkmark$}};
\node[smalllabel,anchor=west] at (2.82,2.42)
  {$2/8$~{\color{green!55!black}$\checkmark$}};
\node[smalllabel,anchor=west] at (2.82,2.10)
  {$3/8$~{\color{green!55!black}$\checkmark$}};
\node[smalllabel,anchor=west] at (2.82,1.78)
  {$8/8$~{\color{red!75!black}$\boldsymbol{\times}$}~{\tiny saturated}};

\draw[line width=0.6pt, green!55!black]
  (3.75,2.10) -- (3.75,2.78);
\draw[line width=0.5pt, green!55!black]
  (3.71,2.78) -- (3.75,2.78)
  (3.71,2.10) -- (3.75,2.10);

\node[font=\sffamily\bfseries\tiny, text=green!50!black, anchor=west,
      align=left] at (3.78,2.42) {$1$--$3@8$\\positives};

\coordinate (labels1west) at (2.80,2.55);

\draw[flow] (base1.east) -- (doc1west);
\draw[flow] (docseast1) -- (acts1.west);
\draw[flow] (docsouth) .. controls ($(docsouth)+(0,-0.6)$)
                          and    ($(solver1.north east)+(0.4,0.4)$) ..
                          (solver1.north east);
\draw[flow] (solver1.east) -- (labels1west);

\begin{scope}[shift={($(solver1.south)+(0,-0.55)$)}]
  \fill[red!80] (-0.20,0) -- (0.20,0) -- (0,0.32) -- cycle;
  \draw[red!90,line width=0.4pt] (-0.20,0) -- (0.20,0) -- (0,0.32) -- cycle;
  \node[font=\sffamily\bfseries\tiny,text=white] at (0,0.12) {!};
\end{scope}
\node[redlabel,anchor=west]
  at ($(solver1.south)+(0.30,-0.39)$) {bottleneck for RL};

\node[font=\sffamily\itshape\scriptsize,align=center,text width=5.20cm]
  at (2.825,0.55)
  {\textbf{Solver label amortized}; solver-in-the-loop RL is prohibitive.};


\node[font=\sffamily\tiny, align=center] (targets) at (8.20,4.45)
  {\textbf{Targets}\\
   $y\!=\!1$~~{\color{green!55!black}$\checkmark$}~$1$--$3@8$\\
   $y\!=\!0$~~{\color{red!75!black}$\boldsymbol{\times}$}};

\node[font=\sffamily, align=center] (data) at (8.20,3.05)
  {$\{(h_i,\,y_i)\}_{i=1}^{N}$\\
   {\tiny Training data}};

\node[probebox, minimum width=1.30cm, minimum height=0.85cm] (probe2) at (8.20,1.75)
  {Probe\\[-0.4mm]{\tiny binary classifier}};

\draw[flow] (targets.south) -- (data.north);
\draw[flow] (data.south)    -- (probe2.north);

\node[font=\sffamily\itshape\scriptsize, align=center, text width=4.40cm]
  at (8.20,0.55)
  {\textbf{Train probe} to predict $y\!\in\!\{0,1\}$ from $h_i$.};



\node[box, minimum width=2.30cm, minimum height=0.85cm] (ref) at (12.25,4.40)
  {Reference Model\\[-0.4mm]{\tiny frozen copy of base}};

\pic at (12.10,3.12) {doc};
\coordinate (tasktop) at (12.25,3.48);
\coordinate (taskbot) at (12.25,3.12);
\node[smalllabel,anchor=west] at (12.55,3.30) {Generated\\task};

\node[box, minimum width=1.60cm, minimum height=0.85cm] (student) at (12.25,2.20)
  {Generator\\[-0.4mm]{\tiny $\pi_\theta$}};

\node[probebox, minimum width=1.20cm, minimum height=0.85cm] (probe3) at (15.00,4.40)
  {Trained\\Probe};

\node[smalllabel] (predlabel) at (15.00,3.20) {Predicted\\difficulty};
\node[anchor=center, font=\sffamily] (pred) at (15.00,2.20) {$0.68$};

\draw[flow]   (student.north) -- (taskbot);
\draw[flow]   (tasktop) -- (ref.south);
\draw[frozen] (ref.east) -- (probe3.west);
\draw[flow]   (probe3.south) -- (predlabel.north);
\draw[flow]   (predlabel.south) -- (pred.north);
\node[smalllabel, anchor=south] at (13.90,4.50) {$h(t)$};

\draw[reward] (pred.west) -- (student.east);
\node[redlabel, anchor=south] at (13.95,2.45) {RL update};

\node[font=\sffamily\itshape\scriptsize, align=center, text width=4.40cm]
  at (13.35,0.55)
  {\textbf{Probe-only reward} -- no solver trials inside the RL loop.};

%

\draw[black!70,line width=0.5pt] (16.65,1.55) -- (18.40,1.55);
\draw[black!70,line width=0.5pt,->] (16.67,1.52) -- (16.67,4.15);

\foreach \cx/\hPre/\hPost in {%
  16.83/2.40/0.30,
  17.18/0.40/1.40,
  17.52/0.40/2.05,
  17.87/0.40/1.30,
  18.22/2.20/0.65%
} {
  \draw[fill=black!20,draw=black!60,line width=0.4pt]
    (\cx-0.13,1.55) rectangle (\cx-0.02,1.55+\hPre);
  \draw[fill=green!55!black!40,draw=green!45!black,line width=0.4pt]
    (\cx+0.02,1.55) rectangle (\cx+0.13,1.55+\hPost);
}

\draw[fill=black!20,draw=black!60,line width=0.4pt]
  (17.20,4.55) rectangle (17.32,4.65);
\node[font=\sffamily\tiny,anchor=west] at (17.36,4.60) {pre-RL};
\draw[fill=green!55!black!40,draw=green!45!black,line width=0.4pt]
  (17.20,4.32) rectangle (17.32,4.42);
\node[font=\sffamily\tiny,anchor=west] at (17.36,4.37) {post-RL};

\foreach \cx/\lbl in {16.83/{0}, 17.18/{1}, 17.52/{2}, 17.87/{3}, 18.22/{$\geq 4$}} {
  \node[font=\sffamily\tiny] at (\cx,1.38) {\lbl};
}

\node[font=\sffamily\bfseries\tiny] at (17.52,1.1)
  {Solver passes (@ $k{=}8$)};

\node[font=\sffamily\itshape\scriptsize, align=center, text width=2.65cm,
      anchor=north]
  at (17.52,0.95)
  {\textbf{Solver eval:} tasks shift toward $1$--$3@8$.};

\end{tikzpicture}
}%
\caption{\textbf{Pipeline overview:} (1) A base generator produces a
one-time pool of tasks that are labeled with an expensive solver.
(2) A probe is trained to predict those difficulty labels from the
generator's hidden states. (3) During RL, the generator proposes
tasks; a frozen reference model produces activations and the trained
probe converts them into reward, so the solver is never invoked inside
the inner loop. (4) The trained generator is finally evaluated against
a held-out solver to confirm that probe-driven shaping translates into
true difficulty gains.}
\label{fig:rlfr}
\end{figure}
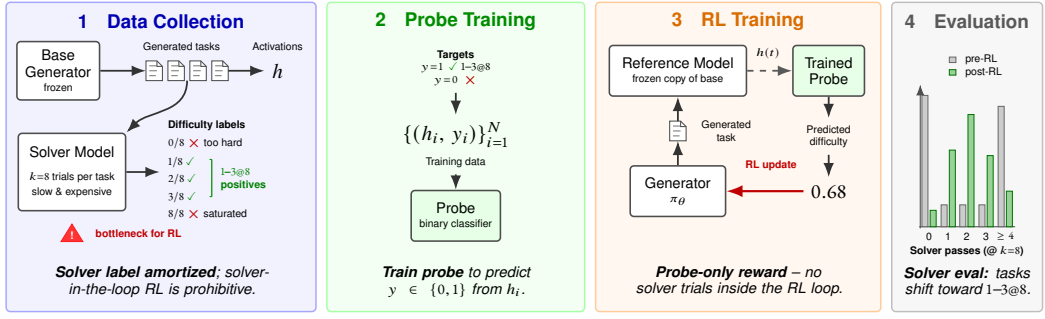

Reinforcement learning on verifiable rewards (RLVR) has become the dominant recipe for eliciting reasoning and agentic behavior from language models \citep{guo2025deepseek,lambert2025reinforcement,liu2025prorl}. Progress under this recipe is gated by the supply of training tasks. As policies improve, fixed task distributions saturate, and further gains require harder tasks that remain discriminative at the current capability frontier. Hand-curated benchmarks cannot keep pace, and naive synthetic generation tends to produce tasks that are either trivially solvable or ill-posed. A natural alternative is to train a \emph{generator} model with RL, rewarding it for producing tasks that are well-formed and appropriately difficult for a target solver \citep{zhao2025absolute,wei2025toward}. The implied objective is \emph{discriminability}, tasks on which the solver is challenged but does not completely fail \citep{wei2025toward}.

However, evaluating a candidate task requires running the solver, and in agentic settings this can be prohibitively expensive. On SWE-bench-style tasks \citep{jimenez2024swebench,Yang2025SWEsmithSD} a single rollout can take tens of minutes since it involves repository navigation, tool calls, and test execution. A reliable difficulty signal needs many such rollouts per candidate to estimate solve rate. Embedding this loop inside generator RL makes training on meaningful distributions infeasible. The same bottleneck appears, more mildly, in competitive math and code generation, where solver trials are cheaper but still costly and variance remains high. Standard RLVR pipelines do not scale to objectives whose verifier is itself an expensive stochastic agent; synthetic generation of tasks scales doubly poorly as solve rate, complexity, and cost-to-solve are unfavorable.

We introduce \textbf{PROPEL} -- \emph{\textbf{P}robe \textbf{R}ewards for \textbf{O}ptimizing \textbf{P}roblems at the \textbf{E}dge of \textbf{L}earning} -- a solver-amortized framework for training task generators. PROPEL builds on Reinforcement Learning from Feature Rewards (RLFR; \citealp{prasad2026features}), which uses interpretability features from hidden states to supervise open-ended generation, and adapts that recipe to task generation with two changes that the agentic setting demands: a multi-step trajectory formulation for software-engineering tasks, and explicit treatment of fixed-probe mode collapse. A small probe is trained once on a one-time labeled corpus of (task, solver-outcome) pairs read out from a frozen reference generator's activations; during RL it replaces live solver rollouts as the reward, collapsing per-step cost from many solver trials to a single forward pass (see Figure~\ref{fig:rlfr}). The construction exploits a well-documented property of language models, that quantities of interest are often represented internally even when the model cannot act on them reliably at generation time \citep{orgad2024llms,zhang2025reasoning}. Provided that a candidate task's well-formedness, solvability, and difficulty-calibration is decodable from generator hidden states, the probe gives a dense, near-free signal that stands in for the true objective long before any solver rollout would confirm it.

\begin{figure}[htbp]
  \centering
  \includegraphics[width=\linewidth]{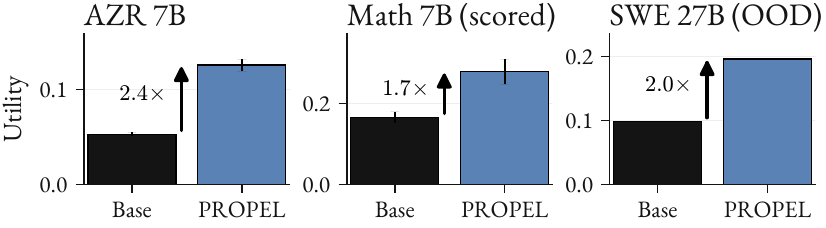}
  \caption{PROPEL significantly outperforms base in terms of utility of generated tasks. Utility is measured based on the \texttt{Qwen2.5-7B-Instruct} solver for AZR and math, and the \texttt{Qwen3.5-27B} solver on held-out OOD repositories for SWE. On math utility is reported on the post-oracle scored tasks. Error bars are $\pm 1$ standard error across RL
  seeds (single seed for SWE). 
  }
  \label{fig:level1-phaseb-headline}
\end{figure}

We show that PROPEL breaks the solver bottleneck in generator training. Training a generator against activation probes rather than live solver trials yields tasks that are harder and more discriminative for the target solver while requiring less than half of the solver trials. PROPEL approximately doubles the rate at which the generator produces learnable-frontier tasks across math, code induction, and software engineering tasks and across solver model sizes (e.g. for code induction $10.1\% \to 20.0\%$, $+98\%$ relative, targeting a \texttt{Qwen2.5-3B-Instruct} solver; $5.3\% \to 12.6\%$, $+138\%$ on \texttt{Qwen2.5-7B-Instruct}, see Figure~\ref{fig:level1-phaseb-headline}). To mitigate the diversity loss that arises when optimizing against a single fixed probe, we apply worst-case optimization (WCO) and adversarial co-evolution of the probe. 
Our contributions are as follows.

\begin{contributions}
   \item \textbf{PROPEL: feature rewards for task generation, including multi-step settings.} We utilize RLFR probe rewards in a solver-amortized generator-RL pipeline that replaces solver-in-the-loop verification with an activation probe, collapsing per-step reward cost from $k$ solver rollouts to a single forward pass. PROPEL makes generator RL tractable in regimes such as agentic SWE, where solver-in-the-loop training is not.\looseness=-1
    \item \textbf{Characterizing mode collapse and mitigating it with worst-case optimization.} We observe mode collapse to a semantic topic under fixed-probe optimization and show that worst-case optimization can mitigate it while maintaining $+86\%$ relative frontier-rate gain over base. We additionally investigate regularization, and adversarial probe co-evolution.
    \item \textbf{Empirical gains across math, code-induction, and SWE at multiple model scales.} On code induction PROPEL approximately doubles the rate at which the generator produces learnable-frontier tasks for the solver ($10.1\%{\to}20.0\%$ on $3$B, $+98\%$ relative; $5.3\%{\to}12.6\%$ on $7$B, $+138\%$). On Math the same recipe shifts the post-strict-oracle conditional yield substantially ($+11$pp for \texttt{Qwen2.5-7B-Instruct}, $+17$pp on \texttt{Qwen2.5-3B-Instruct}). On the significantly more costly and complex SWE domain, PROPEL doubles the rate of learnable-frontier bugs targeting a \texttt{Qwen3.5-27B} solver.
    \item \textbf{Evaluating cold transfer of probes across model families.} We demonstrate cold transfer of a fixed probe across generator families, showing that a probe trained on \texttt{Qwen3.5-4B} drives substantial utility gains when swapped to \texttt{Mistral-7B-Instruct-v0.3} and \texttt{Phi-3.5-mini-instruct} without any per-family retuning, evidence that the encoded utility signal generalizes across model families.
\end{contributions}


\section{Related Work}
\label{sec:related}

\paragraph{Synthetic task generation and self-play.}
A growing body of work trains task generators that target the solver's learnable frontier, with two prior systems most directly informing our design. Absolute Zero~\citep{zhao2025absolute} self-plays a proposer/solver pair across deduction, abduction, and induction tasks; we adopt its induction format and its core observation that the proposer needs a difficulty signal that is neither trivial nor unsolvable. Self-play SWE-RL~\citep{wei2025toward} ports this to bug injection vs.\ repair, with an injection reward that peaks for solver pass rates near the middle of the $0$--$1$ range, directly motivating our solver pass-at-$K$ utility.  Beyond these, task-generation work spans math and symbolic reasoning~\citep{liang2025sws, li2025questa, liu2025saturn, lacombe2025reasoning} and software engineering~\citep{sonwane2025bugpilot, pan2024training, jain2025r2e, xie2026hybrid, zhang2025swe, wang2025swedev, zhu2025training}, with adjacent work on synthetic-data quality~\citep{chen2025synque}, solver-side training~\citep{Da2025AgentRLVRTS}, and abstraction generation~\citep{qu2025rlad}. Of the methods that train a task generator with RL, the reward is generally computed by running the target solver on each candidate; we replace those rollouts with a single forward pass through a probe.

\paragraph{Probes and internal-state rewards.}
Probes have been used to predict reasoning correctness~\citep{zhang2025reasoning, david2025temporal, cencerrado2025no}, score best-of-$N$ candidates~\citep{guo2025mining}, and calibrate judges~\citep{radharapu2025calibrating}. More recently, a series of works have explored using model internals as supervision during training~\citep{zhang2026silence, liang2025clue, prasad2026features}. Most relevant to ours, \citet{prasad2026features} introduce \emph{RL from Feature Rewards} (RLFR), a framework using probes over model internals as scalable reward functions for open-ended tasks. While RLFR isolates features associated with hallucinations, we train a probe to predict a task's training utility (as defined in section \ref{sec:ground-truth}) from the internal features of a task generator. For SWE task generation, this requires us to extend RLFR to multi-turn trajectories.

\paragraph{Reward overoptimization and mode collapse.}
Optimizing any learned reward proxy is subject to Goodhart effects~\citep{gao2022scaling, kwa2024catastrophic, moskovitz2023confronting} and mode collapse, with KL regularization itself capable of driving collapse rather than preventing it~\citep{gxchen2025kl}. Iterated RLHF retrains the proxy on the policy's collapsed outputs~\citep{wolf2025rewardmodeloveroptimisationiterated}, and verbalized sampling preserves diversity at the decoding level~\citep{zhang2025verbalized}. Our adversarial probe-co-evolution follows the iterated-feedback recipe. Probe-high but solver-failed outputs become negatives for the next probe.

\section{Probe Rewards for Optimizing Problems at the Edge of Learning}
\label{sec:method}

We study reinforcement learning for \emph{task-generator} language models across three domains of increasing complexity, math competition tasks, code-induction puzzles (Absolute Zero Reasoner - AZR), and software-engineering tasks (SWE). We first introduce the formal setting then discuss the shared framework PROPEL and provide the per-domain specifics in Section~\ref{sec:domains}.

\subsection{Problem Setting}
\label{sec:setting}

A generator policy $\pi_\theta$ synthesizes tasks $x \sim \pi_\theta(\cdot \mid c)$ given a context $c$ that contains a short instruction, examples, or represents access to a Docker image with a passing test suite. A good task should meet the following two criteria: It should be a valid task, i.e., a math task should have a clear statement, a code-induction task should run, and a bug should apply cleanly and make a previously passing test fail. Second, it should be useful for training: a target solver should sometimes solve it and sometimes fail.\looseness=-1

More formally, the low-cost validity check is a predicate $\mathcal{W} : \mathcal{X} \to \{0,1\}$, where $\mathcal{W}(x)=1$ means that $x$ is syntactically valid and can be graded or executed. The expensive signal is a utility label $U : \mathcal{X} \to \mathbb{R}$, computed from the solver's solve counts, or in the case of SWE, solver rollouts graded against tests. Our objective is to train $\pi_\theta$ to increase $\mathbb{E}_{x \sim \pi_\theta}[U(x)]$ while avoiding expensive calls to $U$ inside the RL loop.

\subsection{Utility Signal for Task Generators}
\label{sec:ground-truth}

The main utility label asks whether a task lies at the target solver's learnable-frontier. For a solver $S$, and $K$ attempts, let $\mu_{S}(x)$ be the mean solve rate for group size of $K$. The utility is defined as:
\begin{equation}
    U_{S}(x) = \mathbb{I}[a \le \mu_{S}(x) \le b].
    \label{eq:utility}
\end{equation}
Following the observation by \citet{wei2025toward}, the optimal solve rate range to target with a group size $K=8$ is $a=1/8$ and $b=3/8$. A task is positive if the solver solves it $1$, $2$, or $3$ times out of $8$. A task solved $0$ times is too hard for that solver, and a task solved $4$ to $8$ times is already too easy. This gives the generator credit for tasks that are learnable but not saturated. For SWE we reduce the number of solver trials to $K=3$ to mitigate the computational burden, and use the utility signal $U_{S}(x) = \mathbb{I}[1/3 \le 
\mu_{S}(x) \le 2/3]$. This slightly broader solve rate band is chosen to account for the small number of tasks generated by the base model that yield exactly one successful attempt and increases the pool of positive samples slightly.

\subsection{Probe-reward RL}
\label{sec:probe-reward}
Evaluating $U$ during RL requires $K$ solver trials per task, which is slow, requires additional memory for solver inference during RL, and each attempt must be verified by a grader, which can be costly, especially in the case of SWE tasks where the grader is derived from a repository's test suite. Instead, we approximate $U$ with an \textbf{activation probe} $f_\phi$. The probe is a small classifier trained on a dataset of generated tasks whose expensive labels have been computed offline. During RL, a newly generated task is given to a frozen copy of the base generator, denoted $\pi_{\mathrm{ref}}$. We read hidden states from this frozen reference model, not from the policy $\pi_\theta$ being updated. For a generated task $x$, the reference model processes the rendered task text or trajectory, and we extract
\begin{equation}
    h_L(x) = \mathrm{Pool}(\mathrm{Hidden}_L(x;\, \pi_{\mathrm{ref}})),
\end{equation}
where $\mathrm{Hidden}_L(x;\, \pi_{\mathrm{ref}})$ are the layer-$L$ hidden states of $\pi_{\mathrm{ref}}$ on input $x$ and $\mathrm{Pool}(\cdot)$ is a pooling rule. The probe maps this vector to either a positive-class probability $p_\phi(x)$ or a raw logit score $\ell_\phi(x)$. The probability is useful for calibrated screening; the logit keeps more dynamic range for RL rewards. Freezing the reference model makes the reward a function of the generated task rather than a function of the changing hidden states of the RL policy.

\paragraph{RL reward signal.} The main reward signal uses validity as a gate and the probe as the ranking signal. Invalid generations receive a fixed penalty $r_{\mathrm{bad}}<0$; valid generations receive the raw probe logit:
\begin{align}
    R_{\text{hard}}(x,\phi) &=
        \begin{cases}
        r_{\mathrm{bad}}, & \mathcal{W}(x)=0,\\
        \ell_\phi(x), & \mathcal{W}(x)=1.
        \end{cases}
\end{align}
The validity predicate decides whether a task can be used at all, while the probe ranks valid tasks by estimated frontier utility. Ablations on the reward signal change only this combination rule. \emph{Probe-only} drops the validity gate and rewards $p_\phi(x)$. \emph{Soft-gated} rewards valid tasks with a clipped probability $\mathrm{clip}(p_{\phi(x)}, 0.1, 0.95)$ instead of a raw logit, while still penalizing invalid tasks. 

Additionally, we investigate an ensemble of probes optimized via worst-case optimization (WCO) to mitigate topic collapse and overfitting to specific features of an individual probe \citep{coste2024reward}:
\begin{equation}
    R_{\text{WCO}}(x,\phi) =
        \begin{cases}
        r_{\mathrm{bad}}, & \mathcal{W}(x)=0,\\
        \min_j \ell_{\phi_j}(x), & \mathcal{W}(x)=1.
        \end{cases}
        \label{eq:wco}
\end{equation}

\paragraph{Probe evaluation.} The trained probes are evaluated based on standard classification metrics (Accuracy, Balanced Accuracy, F1-score) and calibration metrics (ECE). We find that probe accuracy and calibration are not enough to quantify successful RL training. A probe that classifies held-out examples well but assigns almost the same reward to every task sampled from the current policy results in almost no signal during RL. We therefore measure reward variance under the base policy $\mathrm{RVP}(f_\phi, \pi_\mathrm{ref}) = \mathrm{Var}_{x \sim \pi_\mathrm{ref}}\big[R(x,\phi)\big],$ where $R$ is the exact reward value used during RL, including validity gates and ensemble aggregation. RVP is measured before RL on a fixed sample from the base policy. We use it as a selection diagnostic together with calibration, and balanced accuracy. We validate the RVP in Appendix \ref{app:rvp}.

\paragraph{Generator RL.} The generator policy is trained with GRPO for math and code \citep{shao2024deepseekmathpushinglimitsmathematical}, while SWE uses DAPO style advantage calculations with soft long context penalties \citep{yu2025dapo}. All methods use the probe-based reward $R$ for generator optimization. For each context $c$, the policy samples a group of $G$ candidate tasks $\{x_i\}_{i=1}^G \sim \pi_\theta(\cdot \mid c)$. Each candidate is passed through the frozen reference $\pi_{\mathrm{ref}}$ to extract $h_L(x_i)$, from which the probe produces the scalar reward $R(x_i,\phi)$. Implementation details for RL training of the generator are specified in Appendix \ref{app:rl-config}. Because $R$ depends only on the rendered task and the frozen $\pi_{\mathrm{ref}}$, the policy cannot improve its reward by shifting the activations the probe reads~\citep{prasad2026features}. The full pipeline (Figure~\ref{fig:rlfr}) is: collect an offline task pool, compute $U$ for that pool, extract frozen-reference activations, train and select probes, and run RL with $R$.

\section{Evaluation}

\subsection{Metrics}
\label{sec:metrics}
The core downstream metric is the realized utility of fresh generations: we sample tasks from the trained policy, then run the solver $K$ times per task and report the share whose realized solve count lands in the target band defined by $U_{S}$. The probe score is never used for evaluation; every result recomputes $U$ from solver rollouts. We additionally report three text-level diversity diagnostics over all generations per condition: \emph{Self-BLEU-3}~\citep{montahaei2019jointlymeasuringdiversityquality}, \emph{Distinct-3} (unique 3-gram ratio; higher = more diverse), and \emph{top-topic rate} (share of generations in the most common per-domain topic).

\subsection{Domains}
\label{sec:domains}

\paragraph{Math and AZR.}
For both math and code, the generator is \texttt{Qwen3.5-4B}~\citep{qwen35} and we train probes against labels $U_{S}(x) = \mathbb{I}[1/8 \le 
\mu_{S}(x) \le 3/8]$ for two solver sizes, \texttt{Qwen2.5-3B-} \texttt{Instruct}~\citep{qwen2025qwen25technicalreport} and \texttt{Qwen2.5-7B-Instruct}. The two domains differ only in the task format and how the ground truth solution is established.\looseness=-1

In math, the generator is prompted with two in-context examples from GSM8K~\citep{cobbe2021trainingverifierssolvemath}, MATH-500~\citep{hendrycks2021measuringmathematicalproblemsolving}, or AIME-style~\citep{dekoninck2026matharena} pools and writes a competition-style task; a solver attempt is correct when its extracted answer is symbolically equivalent to a verified reference. References use a strict two-oracle policy: \texttt{Qwen2.5-32B-Instruct}~\citep{qwen2025qwen25technicalreport} and \texttt{Phi-4}~\citep{abdin2024phi4technicalreport} each make two attempts, and a task is scored only when all four answers agree.

In AZR we adopt the function-induction format of \citet{zhao2025absolute}: given a function seed, the generator produces a triple $x=(f,\{u_1,\ldots,u_n\},m)$ of a Python function, input tuples, and a natural-language hint; the solver sees five input-output examples plus the hint and must reproduce $f$'s behavior on the remaining 5 hidden inputs. Since the generator commits to executable ground truth, no oracle filter is needed. Validity gates, banned imports, sandboxing details, and the full generator prompts are in Appendix~\ref{app:prompts}.

Both domains are trained with GRPO and LoRA adapters \citep{hu2021loralowrankadaptationlarge}; KL coefficients are  $\beta=0.05$ (AZR) and  $\beta=0.1$ (Math). Full hyperparameters in Appendix~\ref{app:rl-config}.

\paragraph{SWE.}
The SWE setting is a bug-writing task on real Python repositories from the SWE-smith registry~\citep{Yang2025SWEsmithSD}, partitioned into repo-disjoint probe-training, RL-training, and evaluation splits (Appendix~\ref{app:swe-data}). The generator, \texttt{Qwen3.5-27B}~\citep{qwen35}, acts as a multi-turn bash agent with a custom harness (see Appendix \ref{app:swe-bug-gen}) inside a sandboxed repository and must introduce a behavior-changing bug while leaving tests unmodified. The verifier checks that the patch applies, tests are unchanged, the suite still runs, and at least one previously passing test fails. For every valid bug we run $K=3$ solver trials with the same model under the OpenCode harness~\citep{anomalyco2026opencode}, each in a fresh sandbox with the bug-introducing patch produced by the generator applied; a trial is solved iff the solver's final edit passes the verifier suite.

Unlike math and AZR, a SWE task is a full agent trajectory: we extract activations per generated agent turn and aggregate the turn sequence before probing. The probe sweep varies reference-model layer, turn aggregation, probe architecture, and class-imbalance handling, and selects a fixed probe before RL (see Appendix~\ref{app:swe-probe-sweep}).

RL uses SkyRL ~\citep{cao2025skyrl} with vLLM rollout engines~\citep{kwon2023efficient}, and per-repository Docker sandboxes; training rollouts are scored by the repository verifier and the selected activation probe, while the expensive $K=3$ OpenCode~\citep{anomalyco2026opencode} solver labels are reserved for offline probe training and held-out evaluation. The generator is trained for $30$ steps with up to $35$ tool-using turns per trajectory. Invalid generations receive a fixed reward of $-0.2$. Full hyperparameters are in Appendix~\ref{app:rl-config}.\looseness=-1

\section{Probe Training Data and Selection}
\label{sec:probe-training-data}

This section reports the probe-training data and the selected probes used by
PROPEL. The implementation-level sweep,
activation-extraction choices, optimizer settings, and selection details are in
Appendix~\ref{app:probe-training-details}.

\subsection{Probe-training data}
\label{sec:probe-training-sets}

For each (domain, target solver) pair we collect one balanced probe-training set. For SWE the positive class is too sparse to discard data, so we instead train with a class-reweighted loss on all $1{,}757$ rows of the Probe-Train split. The resulting dataset sizes are reported in Table~\ref{tab:probe-training-sizes}; additional details can be found in Appendices~\ref{app:probe-training-details} and~\ref{app:swe-data}.

\begin{table}[h]
\centering
\small
\caption{Probe-training dataset sizes per (domain, target solver). Positives
are tasks in the targeted solve-rate band defined in
Section~\ref{sec:ground-truth}. Math/AZR use balanced datasets; SWE retains all rows and reweights the loss to balance the two classes.}
\label{tab:probe-training-sizes}
\begin{tabular}{lllrrr}
\toprule
Domain & Target solver & Balance & Positives & Negatives & Probe-train rows \\
\midrule
Math & Qwen2.5-3B-Instruct & downsample      & 2{,}340 & 2{,}340 & 4{,}680 \\
Math & Qwen2.5-7B-Instruct & downsample      & 1{,}313 & 1{,}313 & 2{,}626 \\
AZR  & Qwen2.5-3B-Instruct & downsample      & 1{,}765 & 1{,}765 & 3{,}530 \\
AZR  & Qwen2.5-7B-Instruct & downsample      & 1{,}167 & 1{,}167 & 2{,}334 \\
SWE  & Qwen3.5-27B         & weighted loss   & 362 & 1{,}395 & 1{,}757 \\
\bottomrule
\end{tabular}
\end{table}

\begin{table}[h]
\centering
\small
\caption{Probes used as the RL reward, one per (domain, target
solver). Validation balanced accuracy and ECE are computed from the random
80/10/10 split for math and AZR, and from 5-fold cross-repository CV for SWE
(mean across folds, averaged over 3 seeds). All ECE values are pre-calibration; the SWE probe additionally has a post-hoc temperature calibrator applied at RL-scoring time. RVP is the reward variance under the base policy, computed on a fixed sample of $n{=}512$ base-policy completions; SWE RVP is omitted due to computational cost.}
\label{tab:selected-probes}
\begin{tabular}{lllllccc}
\toprule
Domain & Target solver & Layer & Pooling & Arch. & Bal. acc. $\uparrow$ & ECE $\downarrow$ & RVP $\uparrow$  \\
\midrule
Math & Qwen2.5-3B  & $11$    & \texttt{last\_token}      & MLP    & $0.611$ & $0.041$ & $0.008$ \\
Math & Qwen2.5-7B  & $15$    & \texttt{last\_token}      & MLP    & $0.660$ & $0.088$ & $0.033$ \\
AZR  & Qwen2.5-3B  & $11$    & \texttt{mean\_last\_50}   & MLP    & $0.649$ & $0.093$ & $0.145$ \\
AZR  & Qwen2.5-7B  & $11$    & \texttt{mean\_last\_50}   & MLP    & $0.622$ & $0.054$ & $0.116$ \\
SWE  & Qwen3.5-27B & $63$ & \texttt{mean\_last\_half} & linear & $0.594$ & $0.294$ & ---     \\
\bottomrule
\end{tabular}
\end{table}

\subsection{Probe Selection}
\label{sec:probe-selection}
For math and AZR we
sweep reference model layer, token pooling, and linear \emph{vs.} MLP probe
heads. For SWE, where each task is a multi-turn agent trajectory, we also sweep trajectory-level pooling and class-imbalance handling. We only use the selected probe for each (domain, target solver). Probe selection uses held-out balanced accuracy and calibration, with reward variance under policy (RVP) as the final Math/AZR tiebreaker on a fixed sample of $n{=}512$ base-policy completions. For SWE, the held-out split is cross-repository: each fold holds out entire repositories, so the selected
probe must transfer beyond repositories seen during probe training. Table~\ref{tab:selected-probes} lists the selected probes used as the RL reward.

The RVP of the selected Math/AZR probes spans more than an order of
magnitude ($0.008$ to $0.145$). Downstream RL gain (Table~\ref{tab:level1-phaseb-headline}) tracks RVP more closely than validation balanced accuracy: AZR, the highest-RVP domain, shows the largest gain, while math (lowest RVP) shows slightly smaller gains.

\section{Results}
\label{sec:results}

\paragraph{Direct solver-in-the-loop RL is more costly and performs worse.}
To isolate and quantify the solver bottleneck, we train the generator with the ground truth utility signal evaluated during training. This baseline is denoted as solver-in-the-loop (SIL) RL and is run on the AZR domain. The results shown in Table~\ref{tab:solver-in-loop-baseline} confirm that online solver feedback can improve the generator, but at significantly higher cost: training for 30 steps consumes $53{,}664$ solver
trials, and requires a
co-located \texttt{Qwen2.5-3B-Instruct} vLLM solver server during RL. Note that this is in the cheaper AZR domain where solver trials are orders of magnitude faster than in SWE. In contrast, PROPEL uses $22{,}592$ solver trials collected offline to train the probe, then replaces the online solver with the frozen-reference forward pass plus a small probe head. It
makes no solver calls during generator RL, while
achieving a utility improvement over the base model more than twice as large as SIL (Table~\ref{tab:solver-in-loop-baseline}).
\begin{table}[htbp]
\centering\small
\caption{On the AZR domain PROPEL outperforms the more expensive solver-in-the-loop (SIL) baseline reaching higher utility lift over the baseline despite requiring fewer solver trials and less peak memory. PROPEL results are reported for \texttt{Qwen2.5-3B-Instruct} across 3 random seeds with one standard deviation.}
\label{tab:solver-in-loop-baseline}
\setlength{\tabcolsep}{4pt}
\begin{tabular*}{\linewidth}{@{\extracolsep{\fill}}lcccc}
\toprule
Method & Signal cost / task & Solver trials on-/offline & Utility (lift) $\uparrow$ & Valid $\uparrow$ \\
\midrule
SIL & 8 solver trials
  & 53.7k / 0
  & 14.04 (+3.95) & 90.66
  \\
PROPEL & 1 probe eval
  & 0 / 22.6k
  & 19.95 $\pm$ 0.62 (+9.86) & 90.85 $\pm$ 1.89
  \\
\bottomrule
\end{tabular*}
\end{table}

\paragraph{PROPEL shifts generators toward the edge of learning.}

Across the three domains, PROPEL yields significant utility gains over the baseline generator (see Figure~\ref{fig:level1-phaseb-headline} and Table~\ref{tab:level1-phaseb-headline}). On code induction, PROPEL roughly doubles the $U_{S}$ across the output distribution at both solver sizes. On math, PROPEL yields a $1.7\times$ lift at both target sizes on the oracle validated output distribution. The multi-turn SWE setting reproduces these gains at the substantially larger $\texttt{Qwen3.5-27B}$ generator and solver scale and exhibits strong out-of-distribution generalization along two axes. First, the RL training repositories are themselves out-of-distribution for the probe, yet PROPEL still improves utility on them by a clear margin. Second, on a set of 550 tasks generated from a held-out set of 11 repositories never seen during RL, the same checkpoint delivers a $2.0\times$ improvement. Together these results show that the gains transfer beyond the probe's training distribution and beyond the RL training distribution, rather than reflecting overfitting at either stage. This confirms that a notion of solve rate and task difficulty is encoded in the hidden activations of LLMs across domains and at various model scales and that task generators can be consistently optimized to maximize $U_S(\cdot)$ via probe rewards.

\begin{table}[h]
\centering\footnotesize
\caption{Evaluation results on AZR, math, and SWE. Utility is $U_S(\cdot)$ defined in Equation~\ref{eq:utility}. Valid is the rate at which generated tasks pass the validity predicate $\mathcal{W}(\cdot)$ (task specification rules for math, sandbox execution for AZR, bug verification for SWE). Self-BLEU-3 and Distinct-3 measure text diversity; top-topic is the share of the most common semantic topic (top-edited-file rate for SWE). The SWE block reports the \texttt{Qwen3.5-27B} generator on its 11 RL-training repositories and on 11 held-out (OOD) repositories with 50 tasks per repo. Bold marks the within-(domain, solver) winner per metric. Values are mean $\pm$standard error across RL seeds (SWE is a single run).} 
\label{tab:level1-phaseb-headline}
\setlength{\tabcolsep}{4pt}
\resizebox{\textwidth}{!}{%
\begin{tabular}{lcccccc}
\toprule
Method & Solver & Utility $\uparrow$ & Valid $\uparrow$ & Self-BLEU $\downarrow$ & Distinct-3 $\uparrow$ & Top-topic $\downarrow$ \\
\midrule
\multicolumn{7}{l}{\textbf{AZR}} \\
Base & 3B & 0.1009 $\pm$ 0.0046 & 0.6654 $\pm$ 0.0042 & \textbf{0.711 $\pm$ 0.003} & \textbf{0.376 $\pm$ 0.003} & \textbf{0.318 $\pm$ 0.013} \\
PROPEL & 3B & \textbf{0.1995 $\pm$ 0.0062} & \textbf{0.9085 $\pm$ 0.0189} & 0.809 $\pm$ 0.031 & 0.266 $\pm$ 0.036 & 0.735 $\pm$ 0.093 \\
PROPEL WCO & 3B & 0.1725 $\pm$ 0.0038 & 0.8346 $\pm$ 0.0170 & 0.714 $\pm$ 0.020 & 0.360 $\pm$ 0.022 & 0.685 $\pm$ 0.123 \\
Base & 7B & 0.0527 $\pm$ 0.0020 & 0.6748 $\pm$ 0.0049 & 0.720 $\pm$ 0.002 & 0.369 $\pm$ 0.003 & \textbf{0.293 $\pm$ 0.004} \\
PROPEL & 7B & \textbf{0.1255 $\pm$ 0.0063} & \textbf{0.8828 $\pm$ 0.0215} & 0.734 $\pm$ 0.021 & 0.346 $\pm$ 0.025 & 0.674 $\pm$ 0.016 \\
PROPEL WCO & 7B & 0.0981 $\pm$ 0.0034 & 0.8457 $\pm$ 0.0254 & \textbf{0.700 $\pm$ 0.018} & \textbf{0.377 $\pm$ 0.021} & 0.535 $\pm$ 0.057 \\
\midrule
\multicolumn{7}{l}{\textbf{Math}} \\
Base & 3B & 0.2474 $\pm$ 0.0137 & \textbf{0.9663 $\pm$ 0.0054} & \textbf{0.631 $\pm$ 0.000} & \textbf{0.505 $\pm$ 0.004} & \textbf{0.605 $\pm$ 0.008} \\
PROPEL & 3B & \textbf{0.4193 $\pm$ 0.0221} & 0.9604 $\pm$ 0.0103 & 0.708 $\pm$ 0.005 & 0.387 $\pm$ 0.004 & 0.751 $\pm$ 0.015 \\
Base & 7B & 0.1660 $\pm$ 0.0133 & \textbf{0.9663 $\pm$ 0.0054} & \textbf{0.631 $\pm$ 0.000} & \textbf{0.505 $\pm$ 0.004} & 0.605 $\pm$ 0.008 \\
PROPEL & 7B & \textbf{0.2791 $\pm$ 0.0309} & 0.9565 $\pm$ 0.0083 & 0.652 $\pm$ 0.014 & 0.464 $\pm$ 0.016 & \textbf{0.516 $\pm$ 0.044} \\
\midrule
\multicolumn{7}{l}{\textbf{SWE}} \\
Base & 27B & 0.2178 & 0.7769 & \textbf{0.778} & \textbf{0.312} & 0.737 \\
PROPEL & 27B & \textbf{0.2664} & \textbf{0.8042} & 0.882 & 0.226 & \textbf{0.694} \\
Base (OOD) & 27B & 0.0984 & \textbf{0.8356} & \textbf{0.811} & 0.085 & \textbf{0.605} \\
PROPEL (OOD) & 27B & \textbf{0.1959} & 0.7513 & 0.898 & \textbf{0.093} & 0.718 \\
\bottomrule
\end{tabular}
}
\end{table}

\paragraph{Mode collapse and mitigations.}
PROPEL optimizing a single probe achieves the most significant gains in terms of utility, but results in stronger semantic
concentration. On AZR with \texttt{Qwen2.5-3B-Instruct} as the solver, it concentrates approximately $74\%$ of
generated tasks on the single semantic topic \texttt{sorting\_order}
(Table~\ref{tab:level1-phaseb-headline}). While it is expected that the distribution of tasks falling inside the targeted utility band is narrower, we attempt to mitigate the loss of diversity via an Ensemble of two probes \citep{coste2024reward}. PROPEL with worst case optimization (WCO) as defined in Equation~\ref{eq:wco} reduces top-topic concentration on AZR with the \texttt{Qwen2.5-7B-Instruct} solver from $0.67$ to $0.54$ while maintaining significant gains in terms of utility over the base generator. Similarly in the case of the \texttt{Qwen2.5-3B-Instruct} solver, WCO results in a top-topic decrease from $0.74$ to $0.69$ while retaining a $+71\%$ utility increase over the base generator. Additionally, we investigated adversarial probe co-evolution, in which false positives of the probe (valid tasks scored as useful but falling outside the true utility band) are mined as negatives to train an auxiliary probe that then constrains the generator's reward. A small investigation on AZR (Appendix~\ref{app:azr-3b-adversarial-coevolution}) recovers most of the probe utility and validity gains without the accompanying semantic collapse, providing preliminary evidence that this failure mode is mineable rather than terminal.

\begin{figure}[htbp]
    \centering
    \includegraphics[width=\linewidth]{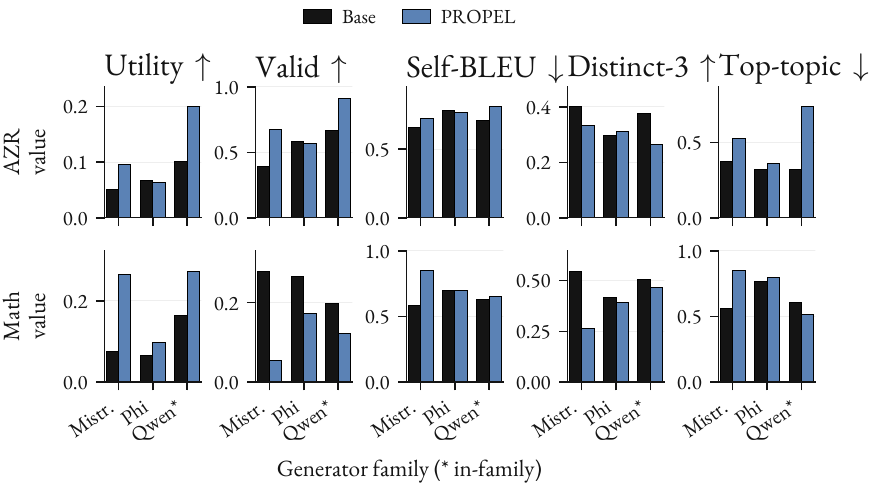}
    \caption{Cross-family probe transfer ($n=1024$ generations per condition). The probe and the \texttt{Qwen3.5-4B} reference model are fixed; only the trainable policy varies. Hyperparameters are inherited from the in-family training runs with no per-family tuning. The \texttt{Qwen}$^*$ (in-family) results initialize the trained policy from the reference model.}
    \label{fig:cross-family-transfer}
\end{figure}

\paragraph{Cold transfer of the probe to different generator families.} The probe and the \texttt{Qwen3.5-4B} reference model are held fixed, while the trainable generator policy is swapped to either \texttt{Mistral-7B-} \texttt{Instruct-v0.3} \citep{jiang2023mistral7b} or \texttt{Phi-3.5-mini-instruct}~\citep{abdin2024phi3technicalreporthighly}, with all other hyperparameters inherited from the in-family RL training on \texttt{Qwen3.5-4B}; that is, we perform no per-family tuning. This is an explicit cold-transfer of the probe, designed
to bound how far the recipe carries without re-tuning. The results are reported in Figure~\ref{fig:cross-family-transfer}. On AZR with the \texttt{Qwen2.5-3B-Instruct} solver, \texttt{Mistral-7B-Instruct-v0.3} shows a clear increase even without tuning: format validity rises from $39.6\%$ to $67.4\%$ and utility from $5.2\%$ to $9.6\%$ after $30$ steps, resulting in a similarly strong increase relative to its base model as the in-family reference. The utility increase when transferring to \texttt{Phi-3.5-mini-instruct} is more modest on AZR. On math, PROPEL on \texttt{Mistral-7B-Instruct-v0.3} yields a significant increase, with utility rising from $7.4\%$ to $26.4\%$, though diversity metrics degrade. The strong transfer of the fixed probe to \texttt{Mistral-7B-Instruct-v0.3} and more modest transfer to \texttt{Phi-3.5-mini-instruct} provide evidence that the encoded utility signal is represented across models and that probe retraining and new data collection may not be required when changing the trainable generator.

\begin{figure}[htbp]
  \centering

  \begin{subfigure}[t]{0.49\linewidth}
    \centering
    \includegraphics[width=\linewidth]{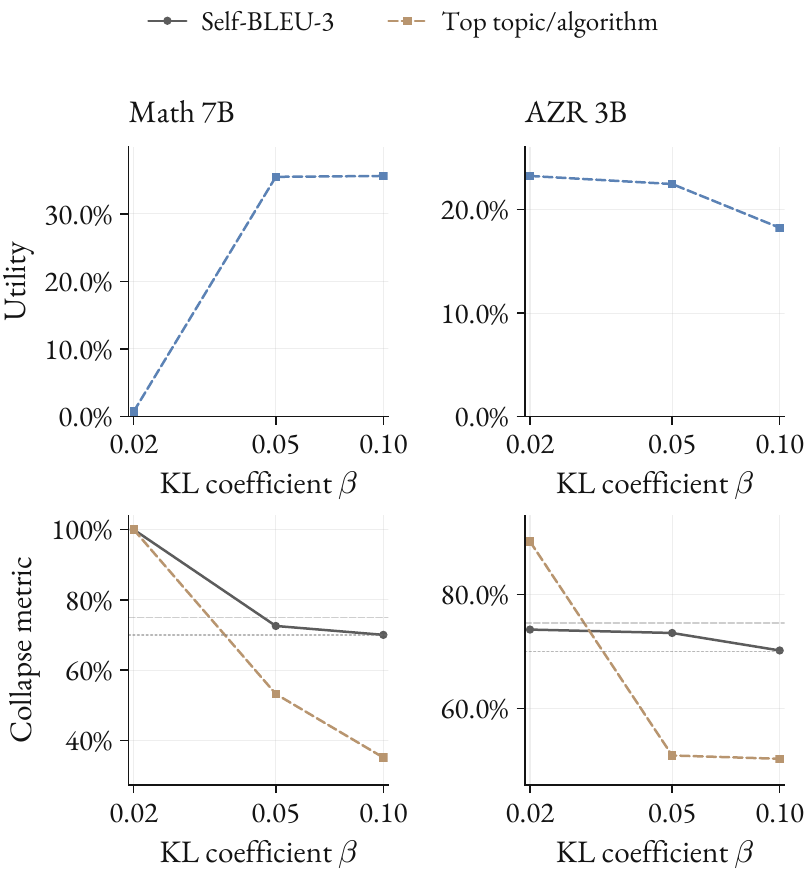}
    \caption{KL ablation.}
    \label{fig:phasea-kl-tradeoff}
  \end{subfigure}
  \hfill
  \begin{subfigure}[t]{0.49\linewidth}
    \centering
    \includegraphics[width=\linewidth]{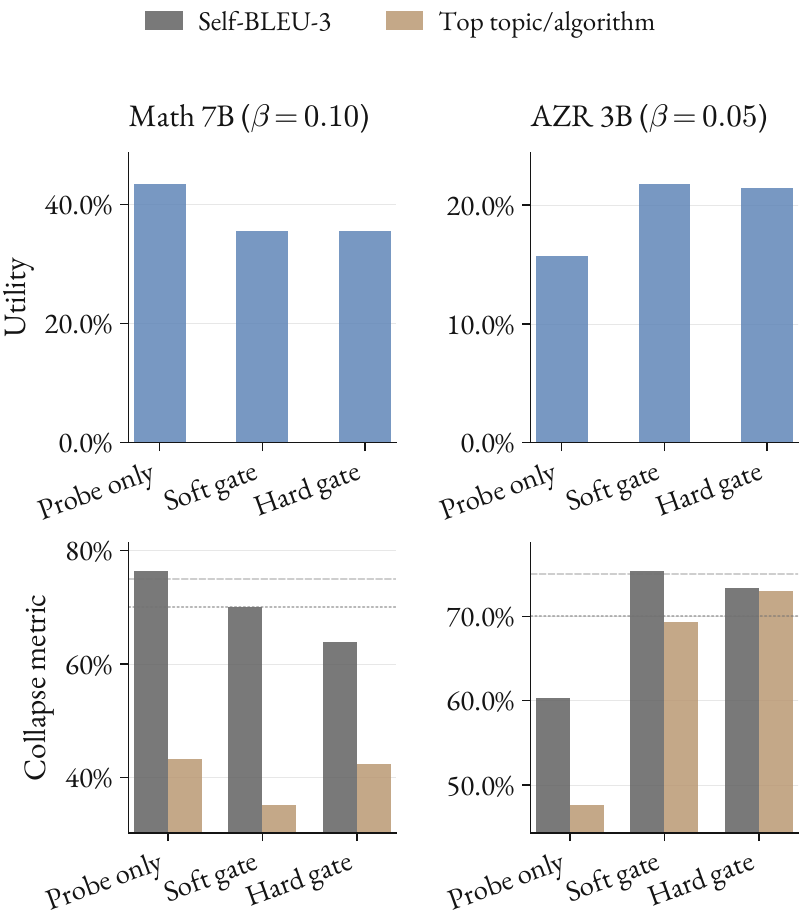}
    \caption{Reward composition ablation.}
    \label{fig:phasea-reward-modes}
  \end{subfigure}

  \caption{Varying the KL regularization strength trades off utility with diversity and topic collapse. The ablation on reward composition shows its impact on the optimization dynamics also depending on the probe.
  }
  \label{fig:phasea-ablation}
\end{figure}

\paragraph{Ablations on KL coefficient and reward composition.}
On math with the \texttt{Qwen2.5-7B-Instruct} solver and on AZR with \texttt{Qwen2.5-3B-Instruct} we ablate the KL regularization strength in Figure~\ref{fig:phasea-kl-tradeoff}. This informed the choice of $\beta=0.1$ for math and $0.05$ for AZR. The results show a clear tradeoff between high utility and low diversity / high topic collapse at low values of $\beta$ with utility and topic collapse decreasing as KL increases. At KL $0.02$ the training run on math had fully collapsed. Next we ablate the reward composition (\emph{probe-only}, \emph{soft-gate}, \emph{hard-gate}) optimized during RL. The results in Figure~\ref{fig:phasea-reward-modes} show that the \emph{hard-gate} variant is consistently the best joint operating point: it matches or beats the others on utility while leaving diversity competitive. 

\section{Conclusion}

\paragraph{Summary.}
We have shown that activation probes trained on a frozen reference model provide a single-forward-pass surrogate for solve rate, and that this surrogate is strong enough to serve as the RL reward for a task generator. Optimizing against the probe shifts generator output toward the learnable frontier of the target solver: on coding tasks the rate at which AZR generators produce frontier-band tasks roughly doubles. Additionally, we provide preliminary evidence that the probe and reference model transfer across generator families, suggesting that the activation signal reflects properties of the task itself rather than of any particular generator.

\paragraph{Limitations.}
Our experiments cover three domains (math, code induction, software engineering); broader coverage of families, scales, and task types is needed before treating these results as general. The probe is trained once on the reference generator's activations and held fixed during RL; under sustained policy drift this surrogate can degrade, and the mode-collapse behavior characterized in Section~\ref{sec:results} is one consequence. Our mitigations (probe ensembling, adversarial co-evolution) are partial and trade efficacy for diversity. Finally, the labeled corpus that anchors the probe still requires solver rollouts to construct; PROPEL amortizes this cost across generator training but does not eliminate it.

\paragraph{Future work.} The most immediate extension is broader empirical coverage: more domains, more generator and solver families, and larger scales where the cost asymmetry between solver rollouts and probe forward passes is more favorable still.

Additionally, throughout this work we treat task generation as a static distributional target: produce tasks near the learnable frontier of a \emph{fixed} solver. In practice, the solver is itself being trained, and the utility of a task depends on what the solver has already mastered. A natural extension is to condition the generator on a trace of recently-attempted tasks and their outcomes so the policy can adapt its target distribution as the solver progresses. Doing so induces an inner-outer loop that is expensive to optimize jointly; how to stabilize it without compounding instabilities across loops is an open question. Using cheap signal from model internals amortizes the dominant cost of this meta-loop (per-iteration solver rollouts) and could provide a tractable path to open-ended learning systems~\citep{hughes2024openendednessessentialartificialsuperhuman}.

Many tasks we ultimately care about (scientific discovery, theorem proving, long-horizon agentic work) have rewards too sparse for direct RL to be effective. A possible extension is to use the generator to propose \emph{goals} rather than full tasks, producing intermediate objectives whose utility decomposes along axes such as achievability, novelty, and relevance~\citep{diazbone2025discover}. 

\paragraph{Outlook.}
Verifiable rewards have carried the current generation of reasoning and agentic models, but the supply of tasks where rewards exist and are cheap and clean is finite. As models saturate existing benchmarks and environments, the bottleneck shifts to producing the right tasks at the right time (a problem that increasingly looks like an RL problem itself). Internal-state rewards offer a way to break the solver-in-the-loop cost barrier that otherwise makes generator RL intractable, and a route into domains where ground-truth verifiers do not exist. 

\bibliography{bib}


\newpage
\beginappendix

\section{SWE RL training results}
\label{app:swe-training-curves}

The selected probe (Table~\ref{tab:app-swe-probe}) drives the
generator-RL run summarized by the \texttt{SWE} block in
Table~\ref{tab:level1-phaseb-headline}; full hyperparameters are listed
in Appendix~\ref{app:rl-config}. Figure~\ref{fig:swe-training-curves}
plots the per-step mean across rollouts of the final reward (validity-gated probe logit with the invalid-reward floor for
non-valid bugs). The reward rises from $0.04$ at the
first step to $0.35$ at the final training step.

\begin{figure}[!htbp]
  \centering
  \includegraphics[width=0.55\linewidth]{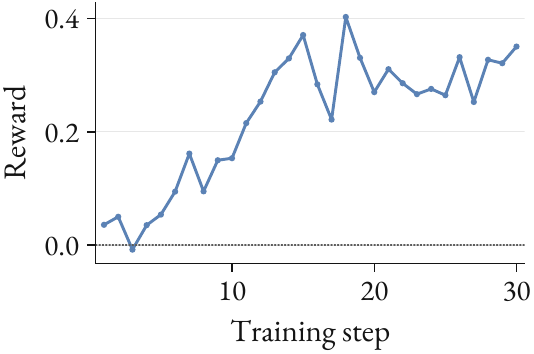}
  \caption{SWE generator-RL training curves: per-step mean of the
  final reward (validity-gated probe logit
  with an invalid-reward floor of $-0.2$). The reward rises from
  $0.04$ at the first step to $0.35$ at the final selected
  checkpoint.}
  \label{fig:swe-training-curves}
\end{figure}

\section{Where the SWE OOD gain comes from}
\label{app:swe-bug-difficulty}

The gain of PROPEL over the base generator on the OOD set is $+9.8$ percentage points, roughly twice the in-distribution gain. The base OOD frontier rate ($9.8\%$) is also well below the base in-distribution rate ($21.8\%$). Both effects are explained by the difficulty distribution of the bugs the generator introduces on unfamiliar repositories. Table~\ref{tab:app-swe-bug-difficulty} breaks down the per-bug solver outcome on the $\geq\!2$-valid-trial subset of each generator. On the OOD set the base generator produces $72.1\%$ of valid bugs in the trivially-solved bucket, versus $61.4\%$ on the in-distribution set. In contrast PROPEL reduces both trivial-solve rates ($18.4\%$ in-dist., $19.6\%$ OOD) and instead concentrates mass at the ``some but not all trials solve'' target band that the probe was trained to reward.

\begin{table}[!htbp]
\centering\small
\caption{Per-bug solver-outcome breakdown by arm, restricted to bugs
with $\geq\!2$ valid solver trials (the denominator used in the
main ``Utility'' column for SWE). \emph{Trivial} ($3/3$ or $2/2$)
and \emph{too hard} ($0/2$, $0/3$) flank the target band ($1$ or $2$
solves out of $\geq\!2$ valid trials).}
\label{tab:app-swe-bug-difficulty}
\begin{tabular}{lcccc}
\toprule
Arm & $n$ ($\geq\!2$ valid) & Trivial & Too hard ($0/k$) & Target band \\
\midrule
Base, in-dist.    & 101 & 62 ($61.4\%$) & 17 ($16.8\%$) & 22 ($21.8\%$) \\
PROPEL, in-dist. & 304 & 56 ($18.4\%$) & 167 ($54.9\%$) & 81 ($26.6\%$) \\
Base, OOD         &  61 & 44 ($72.1\%$) & 11 ($18.0\%$) &  6 ($\phantom{1}9.8\%$) \\
PROPEL, OOD      & 148 & 29 ($19.6\%$) &  90 ($60.8\%$) & 29 ($19.6\%$) \\
\bottomrule
\end{tabular}
\end{table}

Diff size and edit shape are unchanged across regimes: in the $3/3$ subset of the trivial-solve bucket, base on the OOD set produces $13.5$-line diffs (median $13$) with mean $1.2$ inserted and $1.2$ deleted lines, versus $14.3$-line diffs (median $13$) with $1.4 / 1.4$ insertions/deletions on the in-distribution set. The base generator introduces structurally identical one-line edits in both regimes; OOD bugs are just solved more often, consistent with shallow edits landing on less-critical control flow in unfamiliar codebases. The $+9.8$ pp OOD gain therefore reflects RL teaching the generator to land its single-line edit on a behavior-relevant location even when the source repository is novel.

\section{Additional Analysis on SWE}

 \begin{table}[t]
  \centering\small
  \caption{Verifier failure surface for base vs. RL-generated bugs. Rows include only usable bugs under the $\ge 2$ valid-solver-trial filter. Failing tests is the number of verifier tests that fail after applying
  the generated bug patch. Because base and RL are evaluated on the same
  locked repository splits, the absolute failing-test counts are test-suite
  dependent but comparable within each split.}
  \label{tab:failure-surface}
  \begin{tabular}{llrrrrr}
  \toprule
  Split & Generator & Mean & Median & P75 & $\ge 5$ tests & $\ge 10$ tests \\
  \midrule
  RL-Train      & Base &  4.45 & 1 &  5 & 26.7\% & 11.9\% \\
  RL-Train      & RL         & 10.94 & 5 & 11 & 53.9\% & 28.9\% \\
  Held-Out Eval & Base &  5.13 & 2 &  7 & 42.6\% & 18.0\% \\
  Held-Out Eval & RL         & 7.90 & 5 & 11 & 53.4\% & 30.4\% \\
  \bottomrule
  \end{tabular}
  \end{table}

Beyond increasing the number of usable bugs, RL shifts the generated-bug distribution toward broader verifier failure surfaces. On the in-distribution RL-Train split, the median number of failing tests per usable bug increases from $1$ under the base generator to $5$ under the RL generator; the fraction of bugs failing at least 10 tests increases from $11.9$\% to $28.9$\%. On the Held-Out Eval split, the median likewise increases from $2$ to $5$, and the fraction failing at least $10$ tests increases from $18.0$\% to $30.4$\%.

Table~\ref{tab:convergence} shows that mean trajectory length drops from $26.8$ to $15.1$ generator turns over $30$ steps while quality signals rise. Patch-validity climbs from $0.58$ to $0.97$, bug-validity from $0.32$ to $0.79$, and the shaped task reward from $0.28$ to $0.68$. The shift is also visible in the reasons for stopping: clean post-submission terminations (\texttt{stop}) grow from $151$ to $247$ of $256$ episodes, while turn-cap exhaustions (\texttt{max\_turns}) and single-response context overflows (\texttt{length}) collapse from $61$ and $44$ to $2$ and $7$ respectively.

  \begin{table}[t]
    \centering
     \caption{Trajectory length collapses from 26.8 to 15.1 mean turns while
    reward and bug-validity rise. Stop-reason counts are out of 256 episodes per
    step: \textsc{stop} := clean termination after submission, \textsc{max\_turns}
    := hit the 35-turn cap, \textsc{length} := single-response context overflow
    (32k tokens). The episodes shorten because the agent completes tasks
    faster rather than terminating early: \textsc{stop} counts rise while
    \textsc{max\_turns} and \textsc{length} terminations fall.}
    \label{tab:convergence}
    \small
    \setlength{\tabcolsep}{4pt}
    \begin{tabular}{cccccccc}
      \toprule
      Step & turns & Task reward & Bug-valid & Patch-applies & $n(\texttt{stop})$ & $n(\texttt{max\_turns})$ & $n(\texttt{length})$ \\
      \midrule
      1  & 26.8 & 0.276 & 0.324 & 0.582 & 151 & 61 & 44 \\
      10 & 25.6 & 0.432 & 0.492 & 0.633 & 162 & 64 & 30 \\
      15 & 18.7 & 0.661 & 0.777 & 0.914 & 234 &  3 & 19 \\
      20 & 21.0 & 0.577 & 0.673 & 0.836 & 214 & 27 & 15 \\
      30 & 15.1 & 0.676 & 0.788 & 0.965 & 247 &  2 &  7 \\
      \bottomrule
    \end{tabular}
  \end{table}

\section{Solve-count distributions}
\label{app:solve-count-distributions}

\begin{figure}[!htbp]
  \centering
  \includegraphics[width=0.95\linewidth]{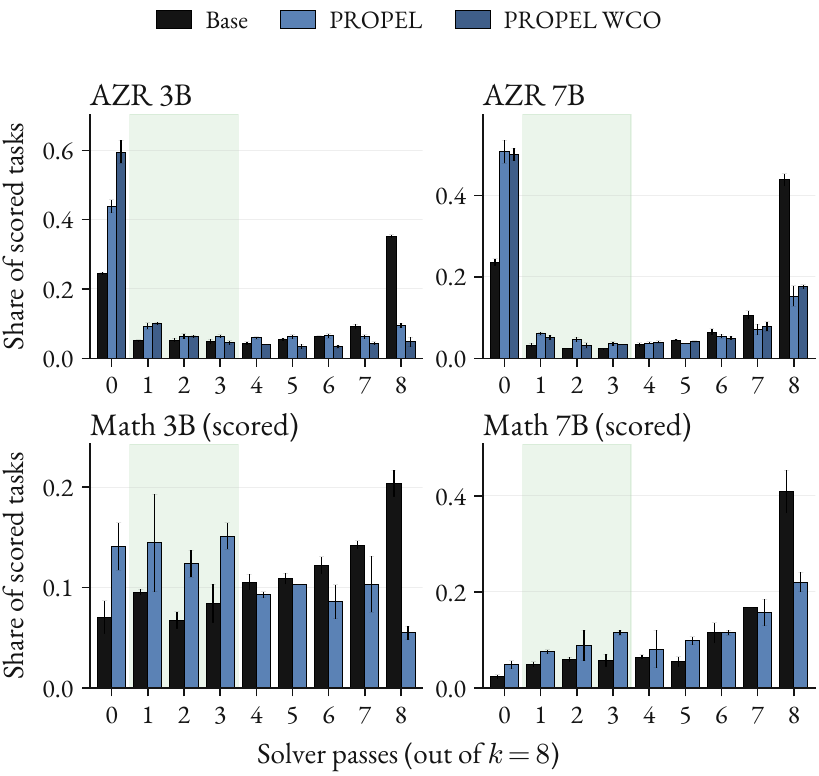}
  \caption{Solve-count distribution PROPEL vs.\ base policy per (domain, target-solver) cell and PROPEL WCO for the AZR domain. The shaded band marks the $1$--$3@8$ band the probe is trained against.}
  \label{fig:solve-count-distribution}
\end{figure}

The subfigures in Figure~\ref{fig:level1-phaseb-headline} report the
aggregate $1$--$3@8$ solve rate per (domain, target-solver) cell. The underlying solve-count distribution is shown in Figure~\ref{fig:solve-count-distribution} (PROPEL and PROPEL WCO vs.\
base policy). The proximal mechanism behind the main lift is the redistribution of mass from the $8/8$ bin (saturated) into the $1$--$3@8$ band the probe is trained against.

\section{AZR 3B adversarial coevolution}
\label{app:azr-3b-adversarial-coevolution}

The AZR single-probe runs exhibit a standard proxy-optimization
failure: the generator learns to produce tasks that score well under
the probe, but many of those tasks come from the same semantic family.
Adversarial co-evolution addresses this by mining the probe's false
positives. A false positive is a valid task that the probe scores as
useful but whose actual target-solver solve count falls outside the
target band. We treat these tasks as negatives, train an auxiliary
probe on them, and fold it back into the next generator-RL run as a
constraint.

Concretely, positive examples are the original probe-training corpus
(valid tasks solved in $1$, $2$, or $3$ of $8$ attempts); negatives
are the false positives just defined plus valid tasks from the
dominant collapsed family discovered after single-probe PROPEL. The
original probe continues to supply the utility signal during RL; the
auxiliary probe acts as a conservative constraint, replacing the
reward on valid tasks with the minimum of the two probe scores.

Using the auxiliary probe in isolation produces a different failure
mode. In a standalone run the generator avoids the original collapsed
family but moves to very easy string-manipulation tasks
($99.7\%$ validity, $1.8\%$ Utility, $88.6\%$ of generations solved
on all $8$ target-solver trials, valid-only top-topic share $99.9\%$).
The auxiliary probe is therefore used as a constraint on top of the
utility objective rather than as a replacement.

Table~\ref{tab:azr3-adversarial-coevolution} reports the follow-up
under the same Utility, Valid, Self-BLEU-3, Distinct-3, and Top-topic
metrics used in the main AZR evaluation. The false-positive variant
raises Utility over base ($0.1009 \to 0.1533$) and Valid over base
($0.6654 \to 0.8398$) while keeping the diversity metrics close to
base, and recovers most of the unconstrained single-probe run's
Utility ($0.1995$) without its semantic concentration.

\begin{table}[h]
\centering\small
\caption{AZR 3B adversarial coevolution as a reward-hacking mitigation. Metrics match Table~\ref{tab:level1-phaseb-headline}: Utility is generated-denominator $1$--$3@8$ frontier rate, Valid is sandbox validity, Self-BLEU-3 is lower-is-better repetition, Distinct-3 is higher-is-better trigram diversity, and Top-topic is the share of generations in the most common semantic family. Base and single-probe rows average three seeds. Adversarial rows are a first follow-up sweep: one RL seed per variant.}
\label{tab:azr3-adversarial-coevolution}
\setlength{\tabcolsep}{4pt}
\begin{tabular}{lccccc}
\toprule
Condition & Utility $\uparrow$ & Valid $\uparrow$ & Self-BLEU-3 $\downarrow$ & Distinct-3 $\uparrow$ & Top-topic $\downarrow$ \\
\midrule
Base & 0.1009 & 0.6654 & 0.7110 & 0.3760 & 0.3180 \\
Single probe & 0.1995 & 0.9085 & 0.8090 & 0.2660 & 0.7350 \\
Semantic adversarial & 0.1372 & 0.8911 & 0.7200 & 0.3670 & 0.4070 \\
False-positive adversarial & 0.1533 & 0.8398 & 0.6930 & 0.3890 & 0.3210 \\
\bottomrule
\end{tabular}
\end{table}

\section{SWE Data Pipeline}
\label{app:swe-data}

\subsection{Repository sourcing}

All repositories used in this work are drawn from the \textbf{SWE-smith} Python
profile registry \citep{Yang2025SWEsmithSD}, which provides for each repository a
designated buggy commit, a pre-built Docker sandbox image, a default test
runner, and the corresponding test framework. From this registry we apply a
four-stage filter to obtain a tractable, non-flaky working set:

\begin{enumerate}
\item \textbf{Hard-gate exclusions.} We exclude three repositories
(\texttt{paramiko}, \texttt{sqlfluff}, \texttt{autograd}) known to cause
network/I/O flakiness, prohibitively long test suites, or environment
build failures.

\item \textbf{Preflight quality filter.} A pre-pass rus every remaining
SWE-smith repository in its sandbox once on the unmodified commit
and records build status, test count, baseline test durnation, and baseline
test failures. Repositories pass if
\texttt{status = "ok"}, $\text{tests} > 0$, and
$\text{baseline\_failures} \le 1$. The held-out RL-Train
and Held-Out Eval selections allow a small number of slower high-coverage
repositories to preserve split breadth.

\item \textbf{Quality ranking.} Surviving repositories are ranked by
\emph{tests-per-entity} (a coverage proxy) and the top-$N$ are taken,
where $N$ is split-specific.

\item \textbf{Empirical yield filter.} After running the bug-generation
agent and the K=3 solver labeling pipeline, repositories with no
trajectories satisfying our usability criteria
(\texttt{bug\_valid} $\ge 1$ and at least one valid solver trial;
see Sections~\ref{app:swe-bug-gen}--\ref{app:swe-solver-labeling}) are
dropped from the labeled dataset.
\end{enumerate}


\subsection{Splits}

We construct three repo-disjoint splits, summarized in Table~\ref{tab:app-splits}.

\begin{table}[!htbp]
\centering\small
\caption{Labeled-dataset splits. Repos and Eps are attempted split totals;
Bugs are usable $K{=}3$ rows under two filters ($\ge 1$ valid solver trial
vs.\ $\ge 2$ valid solver trials); zero-yield repositories remain in the
rollout denominator but contribute no rows. Yield is Bugs ($\ge 2$) / Eps.
Valid \% is trial-level validity (fraction of $K \cdot N$ slots with a
non-null reward); Pooled $\bar r$ is solved trials / valid trials over
non-null slots.}
\label{tab:app-splits}
\resizebox{\textwidth}{!}{%
\begin{tabular}{lccccccc}
\toprule
Split & Repos & Eps & Bugs ($\ge 1$) & Bugs ($\ge 2$) & Yield ($\ge 2$) & Valid \% & Pooled $\bar r$ \\
\midrule
Probe-Train     & 46 & 5{,}892 & 2{,}399 & 1{,}757 & 29.8\% & 72.1 & 0.626 \\
RL-Train        & 11 & \phantom{0,}550 & \phantom{0,}109 & \phantom{0,0}85 & 15.5\% & 71.3 & 0.718 \\
Held-Out Eval   & 11 & \phantom{0,}550 & \phantom{0,0}58 & \phantom{0,0}49 & \phantom{0}8.9\% & 73.4 & 0.660 \\
\midrule
Total attempted & 68 & 6{,}992 & 2{,}566 & 1{,}891 & 27.0\% & --- & --- \\
\bottomrule
\end{tabular}%
}
\end{table}

\paragraph{Why $\ge 2$ valid trials.}
The probe-training filter requires at least two valid solver trials
per bug (out of $K{=}3$). A single trial gives a noisy solve-rate estimate
($\hat r \in \{0, 1\}$, variance $0.25$ at true $r = 0.5$); two trials halve
that variance and allow $\hat r \in \{0, 0.5, 1\}$. The filter retains $73.2\%$
of Probe-Train rows ($1{,}757 / 2{,}399$), $78.0\%$ of RL-Train ($85 / 109$), and
$84.5\%$ of the usable Held-Out Eval rows ($49 / 58$), while removing
the noisiest single-trial bugs from training. We report main results under
both filters but use $\ge 2$ for all probe training and held-out scoring.

The roles align with our experimental protocol: \textbf{Probe-Train} is used
for cross-repository cross-validation when training the probe;
\textbf{RL-Train} provides cross-repository held-out evaluation of probe
quality and is also the repository pool against which the generator is RL-trained;
\textbf{Held-Out Eval} is the strict held-out generalization set for the
post-RL generator and supplies bugs for downstream solver fine-tuning.
Repositories never overlap across splits.

\subsection{Bug generation (generator)}
\label{app:swe-bug-gen}

Bug validity (\texttt{bug\_valid}) follows the criteria in
Section~\ref{sec:method}; the verifier replays the agent's diff via
\texttt{git apply -{}-recount} before running the test suite. We run 50
episodes per repository for RL-Train and Held-Out Eval, and up to 280 episodes per repository for the Probe-Train pool. The
Probe-Train pool is filtered to \texttt{bug\_valid = 1} under the
validity check. The offline bug-generation episodes for the RL-Train and Held-Out Eval
splits use max-turns 25, temperature $0.7$, max-tokens $2048$, top\_p
$0.95$, and top\_k $20$; the RL training rollouts themselves use the
larger 35-turn cap (Table~\ref{tab:app-rl-config}).

We define the end-to-end \emph{yield} of a repository as the fraction
of attempted generator episodes that pass the bug-validity gate and
yield at least two valid solver trials. 
Per-repository yields range
from $0\%$ to $91\%$; the dataset-wide yields are $29.8\%$, $15.5\%$,
and $8.9\%$ for Probe-Train, RL-Train, and the full Held-Out Eval
split respectively.

The generator receives the system and instruction prompt shown in Appendix~\ref{app:prompts}.
The full prompt is bash-only. 
The \texttt{browse\_index} reference in
the prompt body refers to a small helper script the generator can invoke
\emph{from the shell} (e.g.\ \texttt{bash -c "python /workspace/browse\_index.py"}), which enables easier repository navigation.

\subsection{Solver labeling}
\label{app:swe-solver-labeling}

Each of the $K = 3$ OpenCode solver trials (Section~\ref{sec:method})
uses three tools (\texttt{browse\_index}, \texttt{show}, \texttt{edit})
and runs in a fresh Docker sandbox with a $1{,}800$\,s wall-clock budget
and 2 CPU / 4\,GB / 10\,GB resource limits. A trial returns reward
$\in \{0, 1\}$ if the verifier ran (the solver submitted a solution
patch within the budget), else \texttt{null} (timeout, setup failure,
or runtime error). The default sampling parameters used for all label runs are
$T = 0.6$, top\_p $= 0.95$, top\_k $= 20$ (the values
loaded by vLLM from Qwen3.5-27B's \texttt{generation\_config.json}).


\section{Probe Training and Selection Details}
\label{app:probe-training-details}

The main probe-training summary reports the dataset sizes and frozen probes
used by RL. This appendix gives the
implementation details behind those choices: activation extraction, the sweep
grid, optimizer settings, and the selection rule.

\paragraph{Activation extraction and candidate grid.}
Activations are read from the frozen reference model. For math and AZR, each
generated task is a single rendered output and is represented by one pooled
activation vector. For SWE, each task is a multi-turn agent trajectory, so we
extract one activation vector per generated assistant turn and pool across
the trajectory. Math and AZR cover an $8 \times 3 \times 2 = 48$-cell grid
per (domain, solver). SWE covers a larger $5 \times 9 \times 3 \times 2 = 270$-cell
grid, with more trajectory poolings, an additional probe architecture, and
class-imbalance handling.

\begin{table}[h]
\centering
\small
\caption{Probe sweep grid and training configuration. The MLP head uses
$d \!\to\! 512 \!\to\! 128 \!\to\! 2$ with dropout $0.3$. SWE additionally
sweeps trajectory-level pooling, a third (deep-MLP) architecture, and
balance strategy, giving the larger $270$-cell grid.}
\label{tab:app-probe-training-config}
\begin{tabular}{lll}
\toprule
 & Math / AZR & SWE \\
\midrule
Reference model & \texttt{Qwen/Qwen3.5-4B} & \texttt{Qwen/Qwen3.5-27B} \\
Layers $L$ & $\{3,7,11,15,19,23,27,31\}$ & $\{5,15,31,47,63\}$ \\
Poolings & \texttt{last\_token}, \texttt{mean\_full}, & \texttt{last\_token}, \texttt{mean\_full}, \\
& \texttt{mean\_last\_50} & \texttt{mean\_last\_\{50,5,3\}}, \\
& & \texttt{mean\_last\_half}, \texttt{max}, \\
& & \texttt{first\_last\_concat}, \texttt{mean\_std\_concat} \\
Architectures & \texttt{linear}, MLP & \texttt{linear}, MLP, deep MLP \\
Balance & downsample & downsample \emph{or} weighted loss \\
Split & 80/10/10 train/val/test & 5-fold cross-repository \\
Max sequence length & $2{,}048$ (math), $4{,}096$ (AZR) & $32{,}768$ \\
Cells per sweep & $48$ & $270$ \\
\bottomrule
\end{tabular}
\end{table}

\paragraph{Training.}
All probe heads are trained with AdamW at learning rate $10^{-3}$, weight
decay $10^{-4}$, batch size $64$, and up to $50$ epochs with early-stopping
patience $7$. Math/AZR probes are fit on downsampled balanced datasets.
SWE keeps all Probe-Train rows under the weighted-loss setting because the
positive class is sparse.

\paragraph{Selection.}
For each (domain, target solver), we first rank sweep cells by held-out
balanced accuracy subject to calibration. For Math/AZR, we then select the
probe by reward variance under policy (RVP) on a fixed sample of
$n{=}512$ base-policy completions. This two-stage rule reflects the empirical
observation that an accurate but nearly constant on-policy probe gives little
useful gradient signal during RL, while a slightly less accurate probe with
higher RVP can produce a stronger downstream utility
lift.

For SWE, the random 80/10/10 split is replaced by 5-fold cross-repository
validation. Each fold holds out entire repositories, testing whether the probe
transfers beyond repositories seen during probe training rather than merely to
unseen bugs from familiar repositories. The selected SWE probe is the
configuration with the best cross-fold balanced accuracy under weighted-loss
training, averaged over $3$ random seeds. On the cross-repository
held-out fold, the Spearman correlation between probe score and ground-truth
solver solve-rate is $\rho = 0.418$.

\subsection{Probe-Sweep Results AZR/Math}
\label{app:probe-sweep-full-grid}

\begin{figure}[!htbp]
  \centering
  \includegraphics[width=0.95\linewidth]{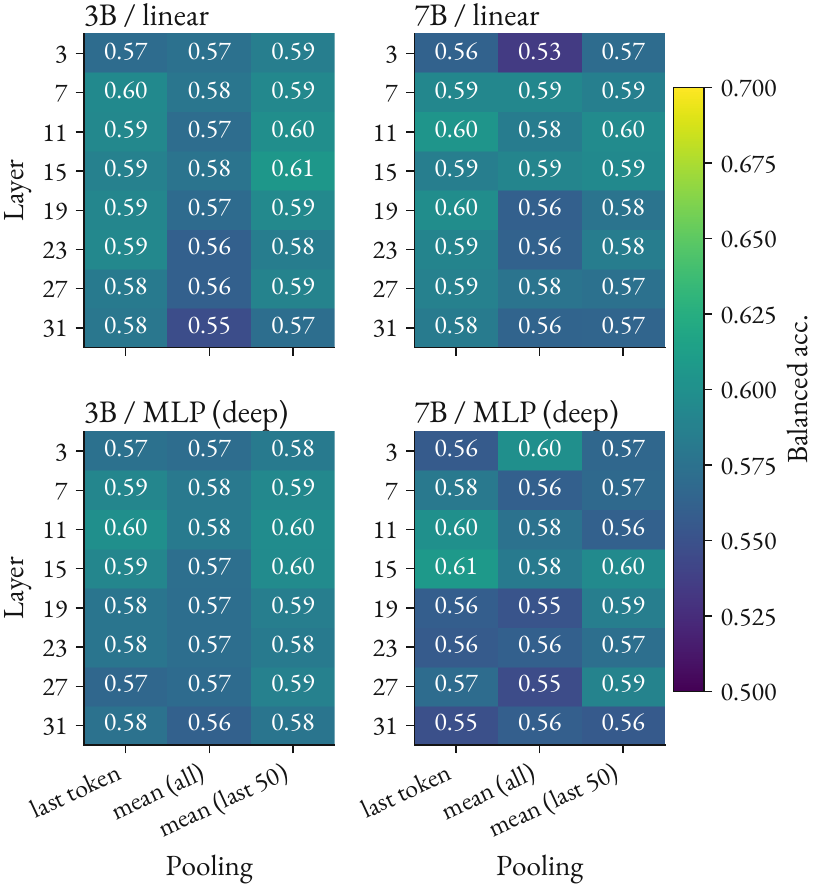}
  \caption{Probe-sweep validation balanced accuracy heatmap on the Math
  domain (1-3@8 label scheme), faceted by architecture (rows) and solver
  size (columns).}
  \label{fig:probe-sweep-heatmap-math}
\end{figure}

\begin{figure}[!htbp]
  \centering
  \includegraphics[width=0.95\linewidth]{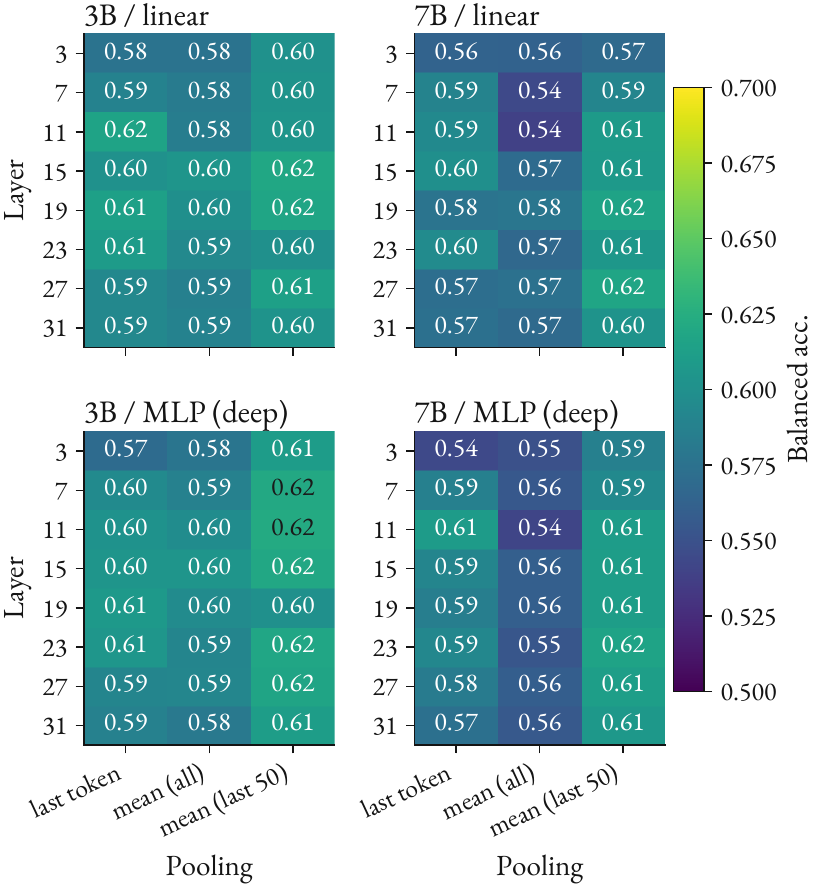}
  \caption{Probe-sweep validation balanced accuracy heatmap on the AZR
  domain (1-3@8 label scheme). Same layout as
  Figure~\ref{fig:probe-sweep-heatmap-math}.}
  \label{fig:probe-sweep-heatmap-azr}
\end{figure}

Figures~\ref{fig:probe-sweep-heatmap-math} and
\ref{fig:probe-sweep-heatmap-azr} render the full $8 \times 3 \times 2$
sweep grids as validation-balanced-accuracy heatmaps for the Math and
AZR domains under the $1$--$3$/8 label scheme, faceted by architecture
(rows) and target-solver size (columns). 

\subsection{Probe Sweep Results SWE}
\label{app:swe-probe-sweep}

We run a complete probe-sweep grid for the SWE domain on the Probe-Train
split (Appendix~\ref{app:swe-data}), training each cell three times with
seeds $\{0, 1, 2\}$ and averaging.

\paragraph{Target.}
The probe is trained on the solver utility signal $U_S$
(Section~\ref{sec:ground-truth}) with $S = \texttt{Qwen3.5-27B}$ at
$K = 3$ solver trials, restricted to bugs with $\ge 2$ valid solver
trials: positives are bugs whose realized solve-rate is in $(0, 1)$
(i.e.\ exactly $1/3$, $2/3$, or $1/2$ when one of three solver trials
is null); negatives are bugs that either fail every valid trial
($0/2$, $0/3$) or succeed on every valid trial ($2/2$, $3/3$). On
Probe-Train this gives $362$ positives and $1{,}395$ negatives
($N = 1{,}757$ across $45$ repositories).

\paragraph{Grid.}
The full $270$-cell SWE sweep ($5\times9\times3\times2$: layer $\times$ trajectory pooling
$\times$ architecture $\times$ class-imbalance strategy) and the shared
training hyperparameters are listed in
Table~\ref{tab:app-probe-training-config}. Epoch selection per cell is
by cross-validated balanced accuracy on the 5-fold cross-repository
split.

\paragraph{Selected configuration.}
Selecting on cross-repo balanced accuracy under weighted-loss training
yields the configuration in Table~\ref{tab:app-swe-probe}.

\begin{table}[!htbp]
\centering\small
\caption{Selected SWE probe configuration, averaged over $3$ random seeds.
Held-out Spearman $\rho$ is computed on the cross-repo held-out fold
(Probe-Train internal CV); CV metrics are mean across the 5 cross-repo folds.
The held-out $\rho$ is high-variance per seed.}
\label{tab:app-swe-probe}
\resizebox{\textwidth}{!}{%
\begin{tabular}{llllcccc}
\toprule
Layer & Aggregation & Arch & Balance & Heldout $\rho$ & CV AUC & CV $F_1$ & CV Bal.\ Acc. \\
\midrule
63 & \texttt{mean\_last\_half} & linear & weighted loss & 0.418 & 0.596 & 0.376 & 0.594 \\
\bottomrule
\end{tabular}%
}
\end{table}


\section{Probe Selection: RVP Validation Panel}
\label{app:rvp}

\begin{figure}[!htbp]
  \centering
  \includegraphics[width=0.6\linewidth]{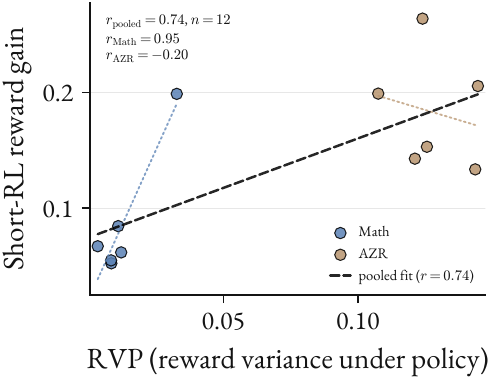}
  \caption{RVP vs short-RL reward gain across the $12$-probe panel.
  Each point is one probe; color marks the domain. Dashed line is
  the OLS fit (pooled Pearson $r = 0.735$, $p = 0.006$, $n = 12$).}
  \label{fig:rvp-vs-gain}
\end{figure}

Section~\ref{sec:method} motivates RVP as a complementary diagnostic
to held-out balanced accuracy and calibration, on the grounds that a
probe with near-constant on-policy reward carries no useful gradient
during RL. We report a $12$-probe panel (6 Math + 6 AZR across two
target-solver sizes), pairing each probe with a matched short-RL run.
RVP is computed once before RL on a fixed sample from
$\pi_{\mathrm{ref}}$. Pooled, RVP and the realized short-RL reward
gain are positively associated (Pearson $r = 0.735$, $p = 0.006$;
Figure~\ref{fig:rvp-vs-gain}), but the effect is largely cross-domain:
within Math the trend is clean ($r = 0.95$, $n = 6$); within AZR the
candidates cluster in a narrow high-RVP band ($0.11$--$0.14$) where
RVP no longer discriminates ($r = -0.20$, $n = 6$). We therefore use
RVP as a coarse sanity check against very low on-policy variance
rather than a fine within-domain selector, alongside balanced
accuracy and calibration when selecting each probe.

\section{RL Hyperparameters}
\label{app:rl-config}

Table~\ref{tab:app-rl-config} provides the generator-RL hyperparameters
for all three domains. Math and AZR share a single training recipe
launched with \texttt{accelerate} on $4$ GPUs per run; SWE uses the SkyRL stack across 3 nodes ($24$ H100s total) with FSDP. The KL coefficient
$\beta$, group size $G$, and selected checkpoint were screened on a per-cell
grid using validity, diversity (Self-BLEU-3, Distinct-3, top-topic), and
collapse alerts. The KL and reward-composition screening sweeps are
reported in Section~\ref{sec:results}
(Figures~\ref{fig:phasea-kl-tradeoff} and \ref{fig:phasea-reward-modes}).

\begin{table}[!htbp]
\centering\footnotesize
\caption{Generator-RL configuration for the confirmatory multi-seed runs.
The same training recipe is used for both standard PROPEL ($R_{\text{hard}}$)
and PROPEL with worst-case ensemble ($R_{\text{WCO}}$) variants; only the reward gate
differs.}
\label{tab:app-rl-config}
\setlength{\tabcolsep}{4pt}
\resizebox{\textwidth}{!}{%
\begin{tabular}{lll}
\toprule
 & Math~/~AZR & SWE \\
\midrule
\multicolumn{3}{l}{\textit{Generator}} \\
Base model & Qwen3.5-4B & Qwen3.5-27B \\
Trainable parameters & LoRA adapters & full fine-tune \\
LoRA rank $r$ & $16$ & --- \\
LoRA $\alpha$ & $32$ & --- \\
LoRA dropout & $0.05$ & --- \\
LoRA target modules & all 7 attn./MLP projections & --- \\
Mixed precision & bf16 & bf16 \\
\midrule
\multicolumn{3}{l}{\textit{Optimizer}} \\
Algorithm & GRPO~\citep{shao2024deepseekmathpushinglimitsmathematical} & DAPO (SkyRL) \\
Optimizer & AdamW & AdamW \\
Learning rate & $5\!\times\!10^{-5}$ & $1\!\times\!10^{-6}$ \\
Weight decay & $0.0$ (default) & $0.01$ \\
Adam $(\beta_1, \beta_2)$ & $(0.9, 0.999)$ & $(0.9, 0.999)$ \\
Max gradient norm & $1.0$ & $1.0$ \\
\midrule
\multicolumn{3}{l}{\textit{Sampling \& rollouts}} \\
Temperature & $0.9$ & $0.7$ \\
Top-$p$ & $0.95$ & $0.95$ \\
Top-$k$ & --- & $20$ \\
Group size $G$ (rollouts/prompt) & $4$ & $16$ \\
Max generation length & $512$ (Math) / $1024$ (AZR) & $2048$ tokens / turn \\
Max input/context length & $4096$ tokens & $32{,}768$ tokens \\
Max multi-turn rollout depth & $1$ (single-shot) & $35$ turns \\
\midrule
\multicolumn{3}{l}{\textit{GRPO loop}} \\
Per-device train batch (generations) & $8$ & $16$ \\
Gradient accumulation steps & $8$ & $1$ \\
Effective rollouts per step & $8 \times 4 \times 8 = 256$ & $16 \times 16 = 256$ \\
PPO inner iterations & $1$ (on-policy) & $1$ (on-policy) \\
\midrule
\multicolumn{3}{l}{\textit{Advantage \& policy-loss}} \\
Advantage estimator & TRL \texttt{GRPOTrainer} & advantage\_estimator: loop \\
Policy loss type & TRL default & Dr.GRPO-style constant-length scaling \\
Importance-ratio clip $\varepsilon$ & TRL default & dual-clip, $\varepsilon_{\text{low}}{=}0.20$, $\varepsilon_{\text{high}}{=}0.28$ \\
KL anchor in loss & yes ($\beta \cdot \mathrm{KL}[\pi_\theta\|\pi_{\mathrm{ref}}]$) & off (\texttt{use\_kl\_loss=false}) \\
KL anchor in reward & no & off (\texttt{use\_kl\_in\_reward=false}) \\
KL coefficient $\beta$ & $0.05$ (AZR), $0.10$ (Math) & --- \\
\midrule
\multicolumn{3}{l}{\textit{Reward shaping}} \\
Reward composition & $R_{\text{hard}}$ / $R_{\text{WCO}}$ & $R_{\text{hard}}$ \\
Invalid-reward floor $r_{\text{bad}}$ & $-0.2$ & $-0.2$ \\
Validity gate $\mathcal{W}(x)$ & sandbox-runnable / oracle (math) & verifier (\texttt{bug\_valid}$\geq$ threshold) \\
Bug-valid threshold (SWE) & --- & $1.0$ \\
Context-overflow reward & --- & $0.0$ \\
Probe gate behavior & always & gated on \texttt{bug\_valid} \\
\midrule
\multicolumn{3}{l}{\textit{Schedule \& selection}} \\
Total GRPO steps & $50$ & $30$ \\
Checkpoint save interval & every $10$ steps & every $5$ steps \\
\midrule
\multicolumn{3}{l}{\textit{Compute}} \\
Probe inference workers & co-located with policy & $4$ Ray actors $\times 2$ GPUs each \\
Max sequence length & $4{,}096$ tokens & $36{,}864$ tokens \\
Parallelism & accelerate DDP, $4$ GPUs & FSDP, 3 nodes $\times 8$ H100s \\
Wall-clock per run & $\approx 6$\,h on $4{\times}$H100 & $\approx 14.5$\,h on $24{\times}$H100 \\
\bottomrule
\end{tabular}%
}
\end{table}

\section{Qualitative examples: base vs RL on AZR}
\label{app:qualitative-azr}

To make the frontier-rate gain in the main results concrete we surface four
representative AZR-induction tasks from the confirmatory multi-seed runs
underlying Table~\ref{tab:level1-phaseb-headline}: two from the base
generator that illustrate the failure modes the probe reward is shaping
\emph{against} (one with $0$ of $8$ hidden tests passed, i.e.\ the
generated function does not match the natural-language message and the
solver cannot recover it; one with $8$ of $8$, a textbook ``square the
input'' that the solver saturates trivially), and two from the
PROPEL trained generator that land in the $1$--$3@8$ learnable-frontier
band the probe targets. Each example is the shortest valid sample in its
solve-count bucket, drawn from the base evaluation run and the
PROPEL checkpoint at step 30.

\paragraph{Base policy --- 0 of 8 solved (unsolvable / mis-specified).}
\begin{promptbox}[title={AZR / 3B target / Base policy / eval seed 12101 / sample 946}]
# Function
def f(n):
    # your function body
    result = []
    while n > 0:
        digit = n % 2
        result.append(digit)
        n //= 2
    return [x for x in reversed(result)]

# Inputs (10)
  1,
  2,
  100,
  1023,
  15,
  17,
  255,
  0,
  64,
  999

# Message
This function converts a non-negative integer into its binary representation as a list of integers (0s and 1s), starting from the most significant bit. If the input is 0, return [0].
\end{promptbox}

\paragraph{Base policy --- 8 of 8 solved (trivial / saturated).}
\begin{promptbox}[title={AZR / 3B target / Base policy / eval seed 12101 / sample 278}]
# Function
def f(n):
    return n * n

# Inputs (10)
  0,
  1,
  2,
  3,
  4,
  5,
  10,
  -3,
  100,
  -1

# Message
The function takes a single number and returns the square of that number. What is the output for the given inputs?
\end{promptbox}

\paragraph{PROPEL --- 2 of 8 solved (learnable frontier, 1-3@8 band).}
\begin{promptbox}[title={AZR / 3B target / PROPEL (step 30) / RL seed 9101 / sample 867}]
# Function
def f(items):
    if not items:
        return 0
    sorted_items = sorted(items)
    prefix_sums = [0] * len(sorted_items)
    prefix_sums[0] = sorted_items[0]
    for i in range(1, len(sorted_items)):
        prefix_sums[i] = prefix_sums[i - 1] + sorted_items[i]
    return prefix_sums[-1]

# Inputs (10)
  [1, 2, 3],
  [10, 20, 30],
  [5, 5, 5],
  [-1, 0, 1],
  [100, 50, 25],
  [3.5, 2.5, 1.5],
  [],
  [7],
  [1, 3, 5, 7, 9],
  [9, 8, 7, 6, 5, 4, 3, 2, 1]

# Message
The function `f` sorts the input list in ascending order. It then computes a cumulative sum (prefix sum) of the sorted elements. The result is the final element of this prefix sum list (i.e., the sum of all sorted elements). If the input list is empty, the function returns 0.
\end{promptbox}

\paragraph{PROPEL --- 2 of 8 solved (learnable frontier, 1-3@8 band).}
\begin{promptbox}[title={AZR / 3B target / PROPEL (step 30) / RL seed 9101 / sample 917}]
# Function
def f(lst):
    if not lst:
        return 0
    sorted_lst = sorted(lst)
    mid = len(sorted_lst) // 2
    left_sum = sum(sorted_lst[:mid])
    right_sum = sum(sorted_lst[mid:])
    if len(sorted_lst) % 2 == 1:
        center = sorted_lst[mid]
        return left_sum + center
    return left_sum + right_sum

# Inputs (10)
  [1, 2, 3],
  [10, 20, 30],
  [5, 5, 5],
  [1, 3, 5, 7, 9],
  [-2, 0, 2],
  [100],
  [],
  [1, 2, 3, 4, 5, 6],
  [99, 88, 77, 66],
  [0, 0, 0, 0, 0]

# Message
The function takes a list of numbers, sorts them, and returns the sum of the first half plus the middle element if the list length is odd. If the list length is even, it returns the sum of the two halves (which are equal in count). If the list is empty, it returns 0.
\end{promptbox}


\section{Generator prompts}
\label{app:prompts}
\subsection{Math domain}
\begin{promptbox}[title={Math task generator: system + user template}]
[system]
You are a creative mathematics problem designer. You output ONLY a problem statement in the exact format below. No solutions, no hints, no teaching notes, no commentary about difficulty or student struggles.

Output format:
## Problem
[Your problem statement here]

Rules:
- Output NOTHING except the ## Problem header and the problem text
- The problem must be from a DIFFERENT area of mathematics each time (geometry, number theory, combinatorics, probability, algebra, calculus, graph theory, logic, etc.)
- The problem must be well-posed, self-contained, and have a definite answer
- The problem should involve a creative twist or unexpected connection

[user]
Here are example problems from different areas of mathematics:

{examples}

Create ONE new problem from a DIFFERENT area of mathematics than the examples above. Remember: output ONLY the problem in the format:

## Problem
[problem statement]

Do NOT include solutions, hints, teaching notes, or any other text.
\end{promptbox}

\subsection{AZR function induction domain}
\begin{promptbox}[title={AZR/induction task generator: system + user template}]
[system]
You are a creative generator of Python induction tasks. You output ONLY the task components below, in the exact format specified. No extra prose, no commentary, no teaching notes.

Output format:

```python
def f(...):
    # your function body
    ...
```

```input
arg1, arg2
```
```input
arg3, arg4
```
... (exactly {num_inputs} `input` blocks)

```message
<natural-language hint to the test subject>
```

Requirements:
- Name the entry function `f`. It must return a value. At least one argument.
- The function must be DETERMINISTIC: no random, time, datetime, I/O, or external state.
- Execution must complete in < 2 seconds of CPU time on all inputs.
- Do NOT import any of these modules: {banned}
- Each `input` block contains comma-separated Python literals -- the args to `f`. Wrap strings in quotes; dicts/lists/tuples are fine.
- Provide EXACTLY {num_inputs} input blocks. Each input must produce a different output.
- The `message` is a natural-language hint describing what `f` does, intended to help a test subject recover the function from only the (input, output) pairs.
- Do NOT include the code snippet inside the message block.

[user]
Example induction tasks (for format reference only -- invent something NEW):

{examples}

Now create ONE NEW induction task. The function must implement a DIFFERENT algorithm than the examples above. Output the code snippet, exactly {num_inputs} input blocks, and a message -- nothing else.
\end{promptbox}

\subsection{SWE domain}

\begin{promptbox}[title={SWE generator: system prompt + instruction}]
[system]
You are a skilled software engineer. You interact with a computer shell to complete coding tasks.

Your response must contain exactly ONE bash code block with ONE command (or commands connected with && or ||).
Include a THOUGHT section before your command where you explain your reasoning.
Format your response as shown:

THOUGHT: Your reasoning and analysis here.

```bash
your_command_here
```

Failure to follow these rules will cause your response to be rejected.

[instruction]
You have a Linux shell at `/workspace`. One command per turn.

Every response must be: THOUGHT (1-2 sentences), then one ```bash block with your command inside it.

THOUGHT: I need to browse the codebase index to find a good target.

```bash
browse_index
```

## Your goal: introduce a subtle bug that causes test failures

You are a skilled developer. Introduce a bug that:
1. Causes existing tests to fail
2. Has exactly one correct fix
3. Is hard to find by reading test error messages alone

Use `browse_index` to explore the codebase. Read test files to understand what's tested.
Find a function with interesting logic and make a subtle change to its behavior.

The hardest bugs to find change WHERE data goes -- which dict key stores a result,
which index slices a string, what a function returns in an edge case. These are hard
because the test error doesn't point directly to the changed line.

Simple condition flips (if X -> if not X) are too easy -- the test error immediately
reveals which branch is wrong. Avoid those.

Be creative. Explore different files and functions each time. The more novel, the better.
\end{promptbox}

\end{document}